\documentclass[11pt]{article}

\usepackage[final]{acl}

\usepackage{times}
\usepackage{latexsym}

\usepackage[T1]{fontenc}

\usepackage[utf8]{inputenc}

\usepackage{microtype}

\usepackage{inconsolata}

\usepackage{graphicx}

\usepackage{amsmath,amsfonts,bm}

\def\eqref#1{equation~\ref{#1}}

\def\1{\bm{1}}

\DeclareMathAlphabet{\mathsfit}{\encodingdefault}{\sfdefault}{m}{sl}
\SetMathAlphabet{\mathsfit}{bold}{\encodingdefault}{\sfdefault}{bx}{n}

\usepackage{hyphenat}
\usepackage{hyperref}
\usepackage{url}
\usepackage{booktabs}
\usepackage{multirow}
\usepackage[dvipsnames]{xcolor}
\usepackage{tabularx}
\usepackage{makecell}
\usepackage{fontawesome5}
\newcommand\blfootnote[1]{\begingroup\renewcommand\thefootnote{}\footnote{#1}\addtocounter{footnote}{-1}\endgroup}
\title{DualEval: Joint Model-Item Calibration for Unified LLM Evaluation}

\author{
\textbf{Aaron J. Li}$^{1}$,
\textbf{Hao Huang}$^{1}$,
\textbf{Youngmin Park}$^{1}$,
\textbf{Yitong Ma}$^{1}$, \\
\textbf{Wei-Lin Chiang}$^{2}$,
\textbf{Li Chen}$^{2}$,
\textbf{Cho-Jui Hsieh}$^{2,3}$,
\textbf{Bin Yu}$^{1}$,
\textbf{Ion Stoica}$^{1,2}$ \\[4pt]
$^{1}$ University of California, Berkeley \quad $^{2}$ Arena \quad $^{3}$ University of California, Los Angeles \\[6pt]
\small \faGlobe~\href{https://dualeval.github.io/}{\texttt{https://dualeval.github.io/}} \\[2pt]
\small \faGithub~\href{https://github.com/aaron-jx-li/DualEval}{\texttt{https://github.com/aaron-jx-li/DualEval}}
}

\begin{document}
\maketitle
\blfootnote{Correspondence: \texttt{aaronjli@berkeley.edu}}
\vspace{-10pt}
\begin{abstract}
Current LLM evaluation relies on two complementary but often disconnected signals: static benchmarks with objective correctness labels and arena-style preference data that better reflect open-ended user interactions. We introduce \textsc{DualEval}, a latent model-item calibration framework that represents models and evaluation items in a shared space, jointly estimating model ability together with item difficulty and sharpness. We apply \textsc{DualEval} across four domains: coding, math, miscellaneous domain-knowledge tasks, and generic everyday user queries. Our evaluation uses 18 frontier LLMs, static benchmark labels, and reward-model scores validated against held-out human preferences for open-ended model responses. Empirically, our framework produces reliable and balanced model rankings, and its learned item-level profiles support downstream applications such as benchmark compression for sample-efficient evaluation and anomaly detection for contamination or outlier analysis. Overall, \textsc{DualEval} unifies static and arena-style evaluation through joint model-item calibration, producing model rankings and item-level diagnostics that support more sample-efficient, interpretable, and auditable evaluation pipelines.
\end{abstract}

\section{Introduction}
LLM evaluation increasingly draws on two complementary signals: static benchmark correctness and human preferences in open-ended arenas. Static benchmarks provide ground-truth labels and standardized grading protocols \citep{hendrycks2020measuring,chen2021evaluating,austin2021program}, but can become saturated or contaminated as models improve \citep{ott2022mapping,magar2022data}. Arena-style evaluations, such as Chatbot Arena \citep{chiang2024chatbot}, better reflect open-ended user interactions through Bradley--Terry-style rankings \citep{bradley1952rank}, but open-ended preference judgments can involve annotator disagreement, subjectivity, and systematic evaluator biases, making them difficult to interpret without additional structure \citep{van2019best,basile2021we,zheng2023judging}. Recent work has begun to connect benchmark and preference-based evaluation \citep{li2024crowdsourced,ni2024mixeval}, but the two settings still provide weakly integrated views of model capability.

Furthermore, both settings are largely model-centric: they aggregate item-level outcomes into overall model scores, treating all items as equally informative. This misses a key insight: ranking signal depends on model-item interaction. Items that are nearly always solved or nearly always failed provide limited separation among the current model pool; the most informative items are those for which small ability differences among evaluated models translate into detectable differences in correctness or preference outcomes. Modeling this item structure can make evaluation more stable, interpretable, and cost-efficient, while also identifying which items are saturated, discriminative, or future-facing.

Toward this end, we propose \textsc{DualEval}, a unified evaluation framework that jointly estimates model abilities and item properties from both static benchmark correctness and arena-style preference signals. Inspired by Item Response Theory (IRT) \citep{ackerman2003using,lord2008statistical}, \textsc{DualEval} places model abilities and item difficulties on a shared latent scale. For static benchmarks, we use a two-parameter logistic IRT model over binary correctness labels. For arena-style data, we use reward-model scores to construct soft pairwise preference targets and fit them through the same latent ability--difficulty structure. This novel joint formulation allows correctness and preference supervision to update shared model and item parameters, yielding not only model rankings but also item-level diagnostic information.

This joint formulation has two key benefits. First, it bridges static and arena evaluation: correctness labels and preference signals can update the same latent model parameters, allowing each source to compensate for the other's weaknesses. Static benchmarks provide clean, controlled supervision, while arena prompts provide open-ended signals closer to realistic user interactions. Second, \textsc{DualEval} turns evaluation data itself into an object of analysis. The learned item parameters characterize each question’s difficulty and sharpness, identify saturated or low-signal items, and support low-cost subsets that preserve the full-data ranking.

We apply \textsc{DualEval} to four domains: coding, math, miscellaneous domain-knowledge tasks, and generic everyday user queries. The first three domains combine static benchmark labels with reward-model scores for open-ended arena responses, while the generic domain uses arena data only. Our experiments cover 18 frontier LLMs from multiple providers. Arena supervision uses our proprietary scalar reward model, developed internally at LM Arena and validated against held-out human preferences; we also verify robustness using the public \texttt{Skywork-Reward-V2-Qwen3-8B} model. Across domains, \textsc{DualEval} reconstructs both evaluation signals, achieving 88--92\% static-label accuracy in the three static-anchored domains and 68--81\% decisive-pair agreement on arena comparisons. Compared with static-only and arena-only baselines, the joint model yields more balanced rankings across evaluation sources and remains robust under reward-model substitution. Its learned item profiles further support diagnostic applications: selecting the top 10\% of items by Fisher information recovers the full-data ranking with Spearman $\rho \geq 0.95$ across all static-anchored domains, while residual-based anomaly detection recovers injected contamination with AUROC $\geq 0.995$ for static-label targets and AUROC $0.93$--$0.98$ for arena-reward targets under a clean-fit reference.

Overall, we reframe LLM evaluation as joint model-item calibration, and our contributions can be summarized as follows:
\begin{itemize}
    \item We propose \textsc{DualEval}, a novel evaluation paradigm that unifies static benchmark correctness and open-ended preference signals on a shared latent model-item scale, enabling mutually informed frontier model evaluation and item diagnostics.
    \item We evaluate \textsc{DualEval} on 18 frontier LLMs across coding, math, miscellaneous domain knowledge, and generic everyday user queries, showing that it produces balanced rankings across signal sources, stable bootstrap estimates, and robustness to reward-model choice in static-anchored domains.
    \item We demonstrate two diagnostic applications: Fisher-information-based benchmark compression, which recovers full-data rankings from small high-signal subsets, and residual-based anomaly detection, which flags (model, item) pairs whose observed outcomes deviate from the fitted model-item expectations.
\end{itemize}
\section{Related Work}
\label{sec:related}

\paragraph{Static and open-ended LLM evaluation.}
Static benchmarks cover NLI and reasoning \citep{bowman2015large, williams2018broad, wang2018glue}, QA \citep{rajpurkar2016squad, yang2018hotpotqa, kwiatkowski2019natural, liang2022holistic}, and math \citep{hendrycks2020measuring, cobbe2021training, hendrycks2021measuring}, but vary substantially in difficulty and discriminability \citep{zhou2026lost, castleman2025rethinking, wang2026metaeval}. Open-ended preference evaluations such as Chatbot Arena \citep{chiang2024chatbot} and MT-Bench \citep{zheng2023judging} capture user-facing quality through aggregate Bradley-Terry rankings without explicit item modeling. Several efforts integrate the two settings \citep{li2024crowdsourced, ni2024mixeval}.

\paragraph{IRT and efficient evaluation.}
Item Response Theory \citep{embretson2013item} has long been used in educational testing and was applied early to NLP for instance-level analysis \citep{lalor2019learning,martinez2019item}. Recent work uses IRT-style modeling for benchmark diagnostics \citep{zhou2026lost, castleman2025rethinking}, label-free evaluation \citep{robertson2025identity}, data-efficient evaluation \citep{liao2025toward, wang2025rethinking}, and adaptive testing \citep{li2025adaptive}. These methods generally operate within a single evaluation source, most often objective correctness labels. \textsc{DualEval} instead couples static correctness with reward-distilled open-ended preferences in a shared latent model-item space.

\section{Method}
\label{sec:method}

In our \textsc{DualEval} framework, each model $i$ is assigned an ability parameter $\theta_i \in \mathbb{R}$, and each task $q$ is assigned a difficulty parameter $b_q \in \mathbb{R}$ and a sharpness parameter $a_q > 0$, parameterized as $a_q=\exp(k_q)$ with $k_q \in \mathbb{R}$. Intuitively, $\theta_i-b_q$ measures how capable model $i$ is relative to task $q$, while $a_q$ is a sharpness parameter that controls how rapidly the predicted success probability changes as model ability moves relative to item difficulty.

\paragraph{Static IRT.}
For static benchmarks with binary correctness labels, we use a standard two-parameter logistic IRT model:
\[
p_{i,q}=P(y_{i,q}=1)=\sigma(a_q(\theta_i-b_q)),
\]
where $\sigma(\cdot)$ is the sigmoid function. Let $\mathcal{S}$ denote the set of observed static labels. We define the static loss as:
\[
\mathcal{L}_{\mathrm{static}}
=
\mathbb{E}_{(i,q)\in \mathcal{S}}
\mathrm{BCE}\!\left(p_{i,q}, y_{i,q}\right).
\]

\paragraph{Reward-Distilled Pairwise IRT.}
For open-ended evaluations, a reward model assigns each model response a scalar score $r_{i,q}$. Our framework accepts any scalar response-quality scorer; the joint formulation provides robustness to RM noise when static correctness labels are available (Appendix~\S\ref{appendix:cross-rm}). We first standardize rewards globally by $z_{i,q} = \frac{r_{i,q}-\bar{r}}{\sigma_r}$, where $\bar{r}$ and $\sigma_r$ are the mean and standard deviation over all reward observations. For each question $q$ and model pair $(i,j)$, we define a soft pairwise preference target:
\[
p^{*}_{ijq}=\sigma(z_{i,q}-z_{j,q}),
\]
which converts reward gaps into calibrated preference strengths: small gaps yield targets near $0.5$, while large gaps yield more confident preferences.

Given the shared IRT parameters, we define the latent success probability of model $i$ on question $q$ as:
\[
p_{i,q}=\sigma\!\left(a_q(\theta_i-b_q)\right).
\]
The predicted preference probability is then:
\[
\hat P(i \succ j \mid q)
=
\sigma\!\left(\gamma (p_{i,q}-p_{j,q})\right),
\]
where $\gamma>0$ is a learned arena-temperature parameter.

\paragraph{Tie and Both-Bad Filtering.}
Not all reward-derived pairs provide useful relative preference signal. We identify ties using the empirical reward-gap distribution: a pair is marked as a tie if $|z_{i,q}-z_{j,q}| < \delta_{\mathrm{tie}}$, where $\delta_{\mathrm{tie}}$ is chosen as the configured percentile of pairwise absolute reward gaps. Ties are excluded from the arena loss.

We also identify ``both-bad'' pairs, where both models receive low standardized rewards with $\max(z_{i,q}, z_{j,q}) < \tau_{\mathrm{bb}}$, where $\tau_{\mathrm{bb}}$ is chosen as the configured percentile of pairwise maximum rewards. Let $\mathcal{A}_{\mathrm{dec}}$ denote the decisive arena pairs, defined as non-tie and non-both-bad pairs, and let $\mathcal{A}_{\mathrm{bb}}$ denote the both-bad, non-tie pairs. We define the full arena loss as:
\[
\begin{aligned}
&\mathcal{L}_{\mathrm{arena}}
=
\lambda_{\mathrm{pref}}
\mathbb{E}_{(i,j,q)\in \mathcal{A}_{\mathrm{dec}}}
\mathrm{BCE}\!\left(
\hat{P}(i \succ j \mid q), p^*_{ijq}
\right) \\
&-\lambda_{\mathrm{bb}}
\mathbb{E}_{(i,j,q)\in \mathcal{A}_{\mathrm{bb}}}
\left[
\log(1-p_{i,q}) + \log(1-p_{j,q})
\right].
\end{aligned}
\]
The first term fits relative preference evidence from decisive pairs. The second term treats both-bad pairs as absolute failure evidence, pushing both models toward low success probability on that question and avoiding conflicting gradients between ``both models failed'' and ``one model was slightly better''.

\paragraph{Joint Objective.}
The static and arena components share the same latent parameters $(\theta,b,k)$. The full optimization objective is:
\begin{equation*}
\begin{aligned}
&\mathcal{L}
=
\lambda_{\mathrm{static}}\mathcal{L}_{\mathrm{static}}
+
\mathcal{L}_{\mathrm{arena}} \\
&+
\lambda_{\mathrm{reg}}
\left(
\|\theta\|_2^2
+\|b\|_2^2
+\|k\|_2^2
+(\log \tilde{\gamma})^2
\right).
\end{aligned}
\end{equation*}

Let $\gamma_0=4$ denote the default arena-temperature scale. We regularize the relative scale $\tilde{\gamma}=\gamma/\gamma_0$ toward one.

For identifiability, after each optimization step we center model abilities to have zero mean and absorb the same shift into item difficulties. This fixes the translation invariance of the latent scale and keeps reported abilities comparable across fits. Final model rankings are obtained by sorting the learned abilities $\theta_i$.
\section{Experiments}
\label{sec:experiments}

In this section, we first validate \textsc{DualEval}'s joint fit and ranking behavior (\S4.1), then study two practical applications enabled by the learned item profiles: sample-efficient benchmark compression (\S4.2) and residual-based anomaly detection (\S4.3).

\paragraph{Experimental Setup.}
We evaluate \textsc{DualEval} on four settings: coding, math, miscellaneous domain-knowledge tasks, and generic everyday user queries. The first three combine challenging static benchmarks with reward-scored open-ended responses; the generic setting uses arena-style data only, serving as an arena-only stress test without static anchors. We select evaluation-relevant arena prompts using LLM judges, with filtering details in Appendix~\S\ref{appendix:inference}. Static benchmarks provide binary correctness labels, while open-ended responses are scored by our proprietary scalar reward model (RM), developed internally at LM Arena and validated against held-out human preferences (Appendix~\S\ref{appendix:rm-verify}). The primary RM uses a \texttt{Qwen-3-32B} backbone with a reward head and was trained on over five million human-preference pairs using a Bradley--Terry objective. To test robustness to reward-model choice, we also repeat the main static-anchored analyses with the public \texttt{Skywork-Reward-V2-Qwen3-8B} RM, which has lower agreement with held-out human preferences (Appendix~\S\ref{appendix:cross-rm}). For each model-question pair, we generate one response under a consistent inference setup (Appendix~\S\ref{appendix:inference}). Table~\ref{tab:data_models} lists the benchmark composition and evaluated models. For agentic coding tasks that require a fixed execution harness, we use the same harness for all models: Terminus2 for TerminalBench~\citep{merrill2026terminal} and mini-SWE-agent v2 for SWE-Bench~\citep{jimenez2024swe}.

\begin{table*}[t]
  \centering
  \small
  \renewcommand{\arraystretch}{1.10}

  \begin{tabularx}{\textwidth}{@{} l X r r @{}}
  \toprule
  \multicolumn{4}{@{}l}{\textbf{(a) Evaluation data composition} \,---\, four domains, each question answered by all 18 models below.} \\
  \midrule
  \textbf{Domain} & \textbf{Static benchmarks (\#Q each)} & \textbf{Arena \#Q} & \textbf{Total} \\
  \midrule
  Coding  & LiveCodeBench v6 (120), MBPP-Plus (80), SWE-Bench-lite (60), TerminalBench-2.0 (40)             & 300 & 600 \\
  Math & AIME 2025 (30), AIME 2026 (30), HLE-Math (60), Olympiad-Math (80)                                & 200 & 400 \\
  Misc & HLE: Bio.+Med.\ (52), Eng.\ (25), Hum.+Soc.+Sci.\ (90), Other (83); SimpleQA (50) & 300 & 600 \\
  Generic & ---                                                                                              & 500 & 500 \\
  \bottomrule
  \end{tabularx}
  
  \vspace{0.6em}

  \begin{tabularx}{\textwidth}{@{} l X l X l X @{}}
  \toprule
  \multicolumn{6}{@{}l}{\textbf{(b) Frontier LLMs evaluated} \,---\, all 18 models applied to every (domain, question) pair.} \\
  \midrule
  OpenAI & GPT-5.5      & Anthropic & Claude Opus 4.7   & Google   & Gemini 3.1 Pro \\
  OpenAI & GPT-5.4      & Anthropic & Claude Opus 4.6   & Google   & Gemini 2.5 Pro \\
  OpenAI & GPT-5.4-mini & Anthropic & Claude Sonnet 4.6 & Google   & Gemini 2.5 Flash \\
  OpenAI & GPT-5-mini   & Anthropic & Claude Haiku 4.5  & xAI      & Grok-4 \\
  OpenAI & GPT-4.1      & DeepSeek  & DeepSeek-V3.2     & Qwen     & Qwen3-Max-Thinking \\
  OpenAI & GPT-4.1-mini & Mistral   & Mistral Large 3   & Meta     & LLaMA-4 Maverick Instruct \\
  \bottomrule
  \end{tabularx}
  
  \caption{Evaluation data composition (top) and frontier LLMs evaluated (bottom). Static benchmarks provide binary correctness labels; arena prompts are filtered from
   LMArena via LLM judges. The generic domain uses arena prompts only. All 18 models respond to every question, producing one response per (model, question) pair. Benchmark source citations and sampling details are provided in Appendix~\S\ref{appendix:inference}.}
  \label{tab:data_models}
  \end{table*}

\subsection{\textsc{DualEval} Results}
\label{sec:results}

\begin{figure*}
    \centering
    \includegraphics[width=\linewidth]{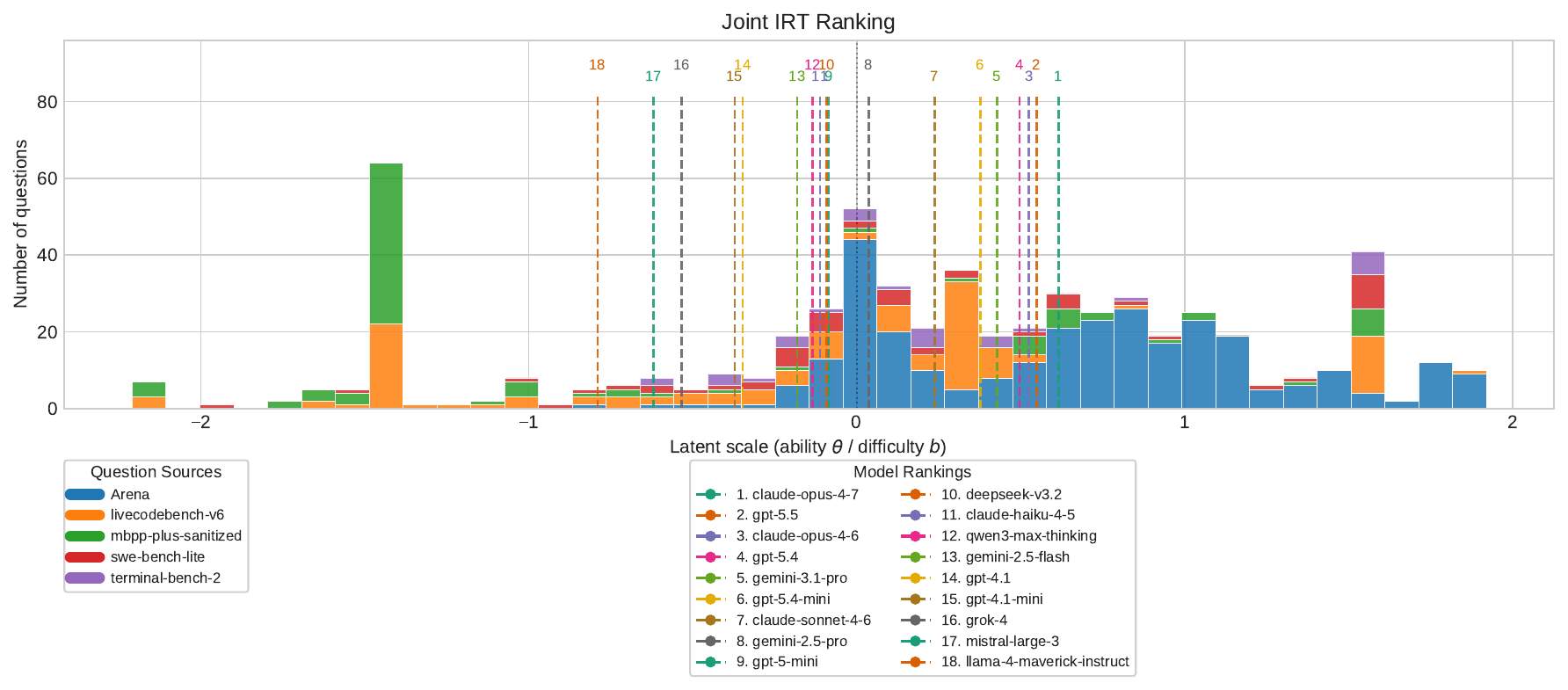}
    \caption{Rankings of model abilities and model-centric question difficulties on the \textbf{coding} domain, using our proprietary reward model.}
    \label{fig:coding_rankings}
\end{figure*}

Figures~\ref{fig:coding_rankings}, \ref{fig:math_rankings}, \ref{fig:misc_rankings}, and \ref{fig:generic_rankings} show the learned model abilities and item-difficulty distributions across the four settings, with math, miscellaneous, and generic visualizations provided in Appendix~\S\ref{appendix:more-rankings}.

\paragraph{Static and Arena Fit.}
\textsc{DualEval} reconstructs both signal types: 88--92\% binary accuracy on static labels across the three static-anchored settings, and 68--81\% decisive-pair agreement on arena comparisons across all four settings. This arena agreement is comparable to the proprietary reward model's agreement with held-out human preferences, suggesting that the joint latent structure captures a substantial share of the preference signal rather than merely fitting noise. Reliability diagrams in Appendix \S\ref{appendix:calibration} show approximate calibration for both static success probabilities and arena preference probabilities, with ECE $0.04$--$0.07$ on static cells and $0.01$--$0.03$ on arena soft pairwise targets, suggesting no systematic biases of over- or under-confidence across domains.

\paragraph{Difficulty Structure.}
The learned item profiles reveal substantial heterogeneity in difficulty and sharpness. Many items lie far from the main ability cluster, indicating saturation or difficulty beyond the evaluated model pool. We also observe a modality difference: static items show near-zero correlation between difficulty $b_q$ and log-sharpness $\log a_q$, whereas arena items show a stronger positive association ($\rho \approx 0.6$ across domains; Appendix~\S\ref{appendix:difficulty_sharpness}). This suggests that difficulty and sharpness contribute differently across evaluation sources, motivating the Fisher-information score in \S\ref{sec:item-informativeness}, which combines both quantities rather than selecting items by difficulty or sharpness alone.

\paragraph{Static--Arena Integration.}
We next test whether joint calibration can integrate static correctness and arena preference signals without collapsing to either source alone. For each static-anchored setting, we compare \textsc{DualEval} with three baselines: a static-only 2PL IRT model trained on benchmark correctness, an arena-only Bradley--Terry model trained on reward-derived pairwise comparisons, and an arena-only average-reward ranking. We average results over three random question splits. Each method is evaluated by held-out arena decisive-pair accuracy and Spearman correlation with two full-data references: static-only IRT and arena-only Bradley--Terry rankings.

Table~\ref{tab:integration} shows the expected source-specialization tradeoff. The static-only model aligns best with the static reference but transfers weakly to the arena reference. Arena-only baselines achieve the highest arena pair accuracy and strongest agreement with the arena reference, but align less well with the static reference. \textsc{DualEval} is designed to balance these sources: it preserves nearly the same static-reference agreement as static-only training while improving arena-reference agreement over the static-only baseline. This supports the role of the joint model as an integration mechanism rather than a single-source optimizer.

We also illustrate the stabilizing role of \textsc{DualEval} when arena supervision is noisier. Replacing our proprietary RM with the public Skywork RM substantially changes arena-only rankings: Arena BT's agreement with the public Arena sub-leaderboards drops to $\rho=0.51/0.54/0.25$ for coding, math, and miscellaneous, respectively. In contrast, \textsc{DualEval} with the public RM preserves high agreement with the static reference ranking, with Static $\rho=0.976/0.984/0.963$, and remains closer to the proprietary-RM \textsc{DualEval} ranking than arena-only training does. A direct cross-RM comparison yields Spearman correlations of $0.963$, $0.992$, and $0.835$ for coding, math, and miscellaneous, respectively (Appendix~\S\ref{appendix:cross-rm}). The arena-only generic setting provides the no-static-anchor counterfactual: without static labels, proprietary- and public-RM rankings diverge sharply ($\rho=0.25$). Together, these results suggest that static anchoring stabilizes the shared latent scale when preference supervision is noisy.

\begin{table*}[t]
\centering
\scriptsize
\setlength{\tabcolsep}{2.5pt}
\begin{tabular}{lcccccccccccc}
\toprule
& \multicolumn{4}{c}{Coding} & \multicolumn{4}{c}{Math} & \multicolumn{4}{c}{Misc} \\
\cmidrule(lr){2-5} \cmidrule(lr){6-9} \cmidrule(lr){10-13}
Method
& Pair Acc. & Static $\rho$ & Arena $\rho$ & Public $\rho$
& Pair Acc. & Static $\rho$ & Arena $\rho$ & Public $\rho$
& Pair Acc. & Static $\rho$ & Arena $\rho$ & Public $\rho$ \\
\midrule
Static 2PL
& 0.731 & 1.000 & 0.773 & 0.763
& 0.641 & 1.000 & 0.571 & 0.682
& 0.701 & 1.000 & 0.695 & 0.701 \\
Arena BT (prop.)
& 0.809 & 0.773 & 1.000 & 0.851
& 0.730 & 0.571 & 1.000 & 0.860
& 0.806 & 0.695 & 1.000 & 0.874 \\
Arena BT (pub.)
& 0.680 & 0.255 & 0.416 & 0.511
& 0.667 & 0.457 & 0.624 & 0.536
& 0.779 & 0.220 & 0.350 & 0.247 \\
Arena Avg Reward (prop.)
& 0.805 & 0.779 & 0.994 & 0.851
& 0.730 & 0.571 & 1.000 & 0.860
& 0.806 & 0.695 & 1.000 & 0.874 \\
Arena Avg Reward (pub.)
& 0.673 & 0.197 & 0.311 & 0.432
& 0.661 & 0.445 & 0.639 & 0.503
& 0.777 & 0.228 & 0.354 & 0.222 \\
DUALEVAL (prop.)
& 0.778 & 0.959 & 0.878 & 0.845
& 0.682 & 0.990 & 0.734 & 0.703
& 0.780 & 0.893 & 0.940 & 0.833 \\
DUALEVAL (pub.)
& 0.591 & 0.976 & 0.805 & 0.789
& 0.600 & 0.984 & 0.625 & 0.692
& 0.612 & 0.963 & 0.724 & 0.657 \\
\bottomrule
\end{tabular}
\caption{
Method comparison across domains and reward models, averaged over three random question splits with a test fraction of 0.2. Pair Acc. is held-out arena decisive-pair accuracy evaluated against soft preference targets derived from the proprietary RM for all methods, including rows trained with the public Skywork RM. All Spearman correlations are computed over the same 18 evaluated models. Static $\rho$ compares the learned model ranking with the full-data static-only IRT ranking; Arena $\rho$ compares it with the full-data Arena BT ranking under the proprietary RM; \textit{public $\rho$} compares it with the matched public LMArena sub-leaderboard over the 18 overlapping models. Coding uses Text-Coding, Math uses Text-Math, and Misc uses Text-Expert. prop./pub. denote the proprietary RM and \texttt{Skywork-Reward-V2-Qwen3-8B}, respectively.
}
\label{tab:integration}
\end{table*}

\paragraph{Ablations.}
A pair-treatment ablation (Appendix \S\ref{appendix:pair-treatment-ablation}, Table~\ref{tab:bb_tie_ablation}) confirms the importance of both-bad anchoring: removing the loss component drops both-bad AUC from 0.903 to 0.622 while leaving held-out arena pair accuracy near 0.78. Tie filtering has a smaller effect on pair accuracy but slightly improves both-bad calibration. The full configuration is used in all subsequent experiments.

\begin{figure*}[t]
\centering
\includegraphics[width=\linewidth]{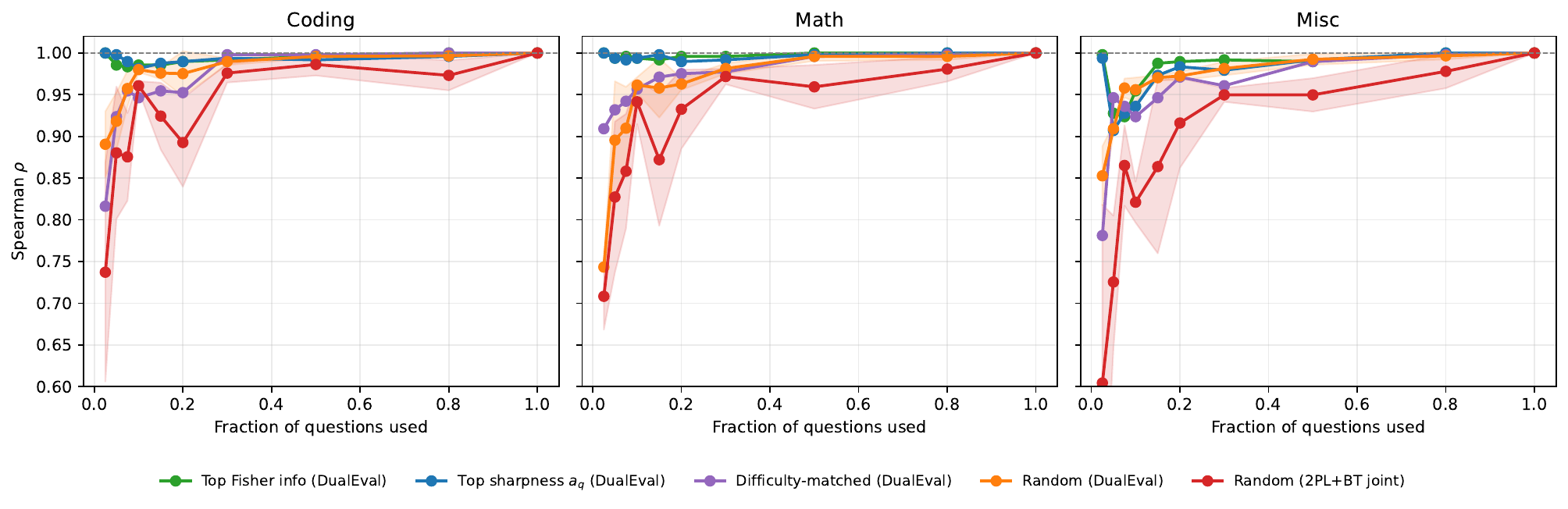}
\caption{Subset-recovery agreement (Spearman $\rho$ vs.\ the full-data \textsc{DualEval} ranking) as a function of the fraction of items retained, across coding, math, and miscellaneous domains. Items are selected by Top Fisher info (\textsc{DualEval}), the sharpness-only and difficulty-matched ablations, and two random baselines. Top
Fisher info recovers the full ranking at $\rho \geq 0.95$ with as little as 10\% of items in all three domains, indicating that ranking signal in current benchmarks is concentrated in a small subset of high-informativeness items.}
\label{fig:sample-efficiency}
\end{figure*}

\subsection{Item Informativeness}
\label{sec:item-informativeness}

A central diagnostic question for any benchmark is which items carry the bulk of the ranking signal between models, and which are redundant. The joint formulation produces item-level parameters that let us answer this question directly: we score each item by its Fisher information about model abilities under the fitted \textsc{DualEval} objective, then quantify how much of the full-data ranking can be recovered from the most informative items alone.

\paragraph{Quantifying Informativeness via Fisher Score.}
For a static item, $p_{i,q}=\sigma(a_q(\theta_i-b_q))$, where $a_q$, $b_q$, and $\theta_i$ are the learned item sharpness, difficulty, and model ability. The Fisher information for model $i$ on item $q$ is $\mathcal{I}^{\mathrm{static}}_{i,q}=a_q^2\,p_{i,q}(1-p_{i,q})$, which we average over models and weight by the static loss coefficient:
\begin{equation*}
S_q^{\mathrm{static}}=
\lambda_{\mathrm{static}}\frac{1}{M}\sum_{i=1}^{M}a_q^2\,p_{i,q}(1-p_{i,q}).
\end{equation*}

For an arena item, \textsc{DualEval} compares responses through $\mu_{ijq}=\sigma(\gamma(p_{i,q}-p_{j,q}))$. We score each pair by the trace of the Fisher information with respect to the two involved ability parameters $(\theta_i,\theta_j)$, which measures local sensitivity of the predicted preference probability to either model's ability:
\begin{equation*}
\begin{aligned}
&\mathcal{I}^{\mathrm{arena}}_{ijq}
=
\mu_{ijq}(1-\mu_{ijq})\gamma^2 \\
&\quad\cdot
\Bigl[
\bigl(a_q p_{i,q}(1-p_{i,q})\bigr)^2
+
\bigl(a_q p_{j,q}(1-p_{j,q})\bigr)^2
\Bigr].
\end{aligned}
\end{equation*}
We average this quantity over decisive arena pairs for item $q$:
\begin{equation*}
S_q^{\mathrm{arena}}
=
\lambda_{\mathrm{pref}}
\frac{1}{|\mathcal{P}_q|}
\sum_{(i,j)\in\mathcal{P}_q}
\mathcal{I}^{\mathrm{arena}}_{ijq},
\end{equation*}
where $\mathcal{P}_q$ denotes the set of decisive, i.e., non-tie and non-both-bad, model pairs for item $q$.

Static and arena items are scored in the same units: expected information about the model ability vector. Averaging over arena pairs prevents items from receiving an artificial advantage from $O(M^2)$ pairwise comparisons.

\paragraph{Subset-Recovery Experiment.}
For each domain, we select a fraction of items, refit ability
parameters on the subset (item parameters held fixed at full-data values), and measure Spearman correlation between the resulting ranking and the full-data \textsc{DualEval} ranking. We compare five subset-selection strategies, including ablations that isolate the contribution of each component of $S_q$: (i) \textbf{Top Fisher info} (\textsc{DualEval}), selecting items by
descending $S_q$; (ii) \textbf{Top sharpness $a_q$}, ablating the $p(1-p)$ frontier weight; (iii) \textbf{Difficulty-matched}, selecting by smallest
$|b_q-\mathrm{median}(\theta)|$, isolating frontier-matched difficulty alone; (iv) \textbf{Random} (\textsc{DualEval}), a uniform random subset; (v) \textbf{Random} (2PL+BT), a uniform random subset refit with a single-source 2PL+BT baseline that lacks item parameters for arena
data.

\paragraph{Ranking Signal Is Concentrated in a Small Subset.}
Figure~\ref{fig:sample-efficiency} reveals substantial item-level redundancy in current benchmarks. Across all three static-anchored domains, the top 10\% of items by Fisher information recover the full-data ranking at Spearman $\rho \geq 0.95$, and the top 30\% recover it at $\rho \geq 0.99$. The full Fisher score is close to the sharpness-only ablation but is the most consistent at small fractions, suggesting that sharpness explains much of the signal while local non-saturation still improves item selection. Difficulty matching alone is less reliable, confirming that proximity to the model-ability range is not sufficient without sensitivity to model differences. The 2PL+BT baseline trails the \textsc{DualEval} variants throughout, indicating that item parameters fit jointly across static and arena data carry information missing from disjoint-source rankings.

\paragraph{Implications.}
This redundancy structure has two practical consequences. First, leaderboard refreshes against established benchmarks can target the high-Fisher subset for substantial reductions in evaluation cost while preserving ranking confidence. Second, the items with lowest Fisher information can be retired without affecting ranking quality, providing a principled diagnostic for identifying saturated or non-discriminative items. Because informativeness is relative to the evaluated model pool, high-signal subsets should be recalibrated as model abilities shift.

\begin{figure*}[t]
\centering
\includegraphics[width=\linewidth]{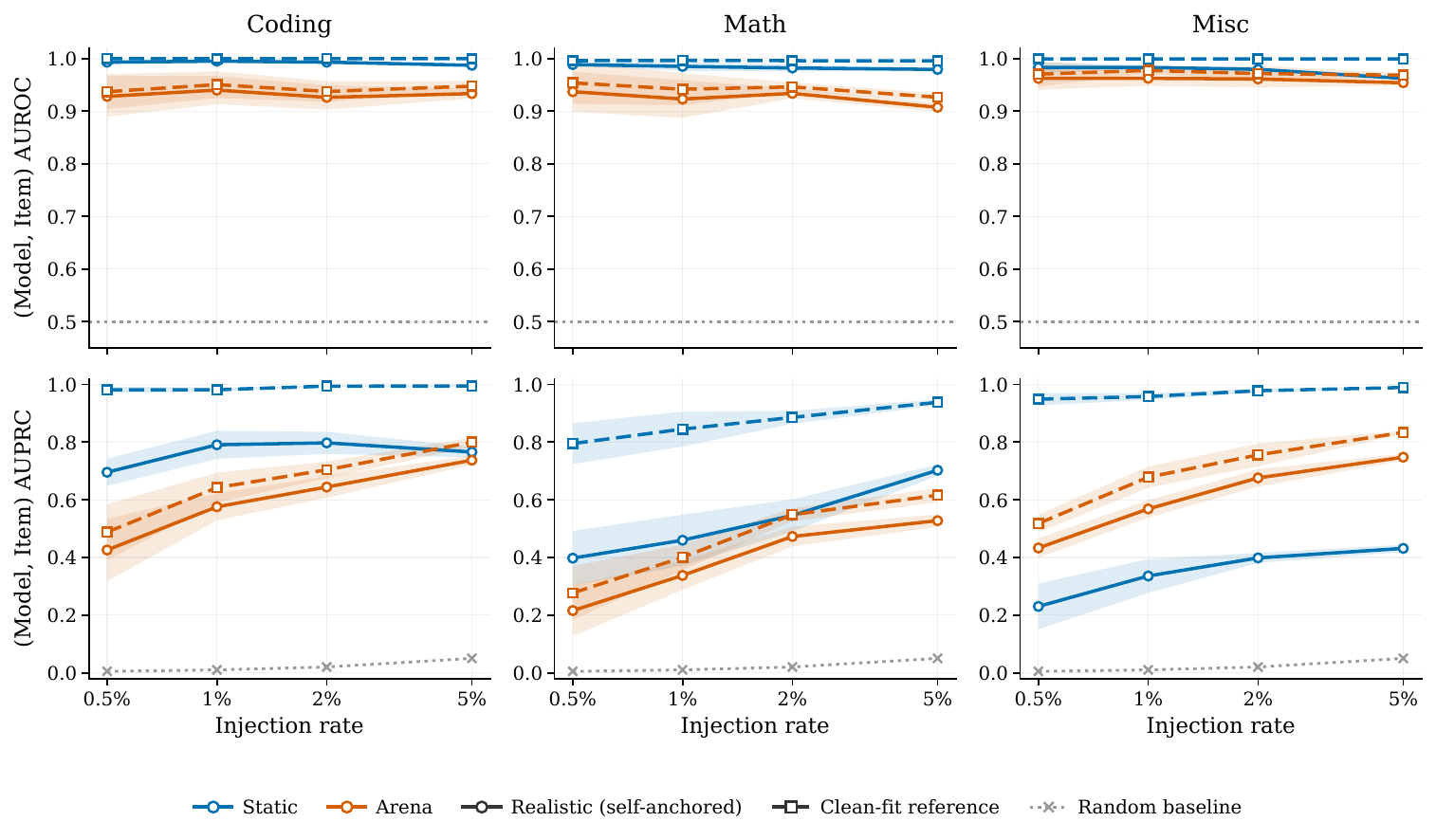}
\caption{Synthetic contamination detection across coding, math, and miscellaneous domains, separately for static-label and arena-reward injections. Each panel plots item-level AUROC/AUPRC versus injection rate; the positive class is the set of synthetically contaminated cells. Residual scores from the clean-reference detector are strongest across domains and injection rates, while the contaminated-fit detector remains above the random baseline.}
\label{fig:contamination}
\end{figure*}

\subsection{Residual-Based Anomaly Detection}
\label{sec:anomaly-detection}
A second diagnostic question is whether the joint fit can identify suspicious $(\text{model}, \text{item})$ pairs whose observed outcomes are inconsistent with what \textsc{DualEval} expects given the model's ability and the item's difficulty. Such residual anomalies are candidates for contamination, leakage, or data-quality issues, and the joint model's per-pair expectations make them directly computable.

\paragraph{Per-Pair Residual Scores.}
For a static $(i, q)$ pair we define the standardized residual
\begin{equation*}
r^{\mathrm{static}}_{i,q} = \frac{y_{i,q}-p_{i,q}}{\sqrt{p_{i,q}(1-p_{i,q})+\varepsilon}},
\label{eq:residual-static}
\end{equation*}
where $y_{i,q}\in\{0,1\}$ is the observed correctness label and $p_{i,q}$ is a
\textsc{DualEval} success probability. Large positive residuals indicate unexpected
successes; large negative residuals, unexpected failures. For an arena $(i,q)$ pair, we average the target--prediction gap across decisive opponents of model $i$ on item $q$:
\begin{equation*}
r^{\mathrm{arena}}_{i,q} = \frac{1}{|\mathcal{P}_q^{(i)}|}\sum_{j\in\mathcal{P}_q^{(i)}}\bigl(\sigma(z_{i,q}-z_{j,q})-\mu_{ijq}\bigr),
\label{eq:residual-arena}
\end{equation*}
where $z_{i,q}$ is the standardized reward, $\mu_{ijq}=\sigma(\gamma(p_{i,q}-p_{j,q}))$ is
the \textsc{DualEval}-predicted pairwise probability, and $\mathcal{P}_q^{(i)}$ is the set
of decisive opponents of model $i$ on item $q$.

\paragraph{Synthetic Contamination Protocol.}
We inject synthetic contamination into existing evaluation artifacts. For each static-anchored domain, we sample candidate cells from low-probability success regions. For static labels, the candidate pool consists of failed cells $(y_{i,q}=0)$ with the lowest $\mathrm{acc}^{\mathrm{model}}_i \times \mathrm{acc}^{\mathrm{item}}_q$, where $\mathrm{acc}^{\mathrm{model}}_i$ is model $i$'s average static accuracy in the domain and $\mathrm{acc}^{\mathrm{item}}_q$ is the fraction of models that solve item $q$. Flipping these cells creates unexpected successes by weak models on hard items. For arena responses, the candidate pool is the bottom $30\%$ of cells by standardized reward $z_{i,q}$. We sample cells uniformly from each pool at rates $\{0.5\%,1\%,2\%,5\%\}$, with five random seeds per rate. Static contamination flips the correctness label from $0$ to $1$; arena contamination adds $2.0$ to the standardized reward $z_{i,q}$ before pairwise targets are rebuilt.

We evaluate two detectors. The contaminated-fit detector scores each cell by its signed positive residual under a \textsc{DualEval} model refit on the contaminated data. The clean-reference detector computes the same scores using a \textsc{DualEval} model fit once on the original clean data, while evaluating the contaminated labels or rewards against clean-fit expectations. Injected cells are positives and all other cells are negatives for cell-level AUROC and AUPRC; Appendix~\S\ref{appendix:contamination-metrics} details how these detection metrics are computed.

\paragraph{Calibrated Residuals Recover Injected Contamination.}
Figure~\ref{fig:contamination} reports AUROC and AUPRC across injection rates and domains. The clean-reference detector recovers injected static-label contamination nearly perfectly, with AUROC $\geq 0.995$ and AUPRC $0.80$--$0.99$, and remains strong on arena-reward contamination, with AUROC $0.93$--$0.98$ and AUPRC $0.28$--$0.83$. This setting is operationally realistic when a benchmark already has a vetted historical calibration and new submissions or data refreshes are screened against it. The contaminated-fit detector, which uses only the corrupted snapshot, remains above the random baseline across all conditions but is less sensitive when flipped labels are absorbed into the refitted latent parameters. The gap between the two detectors highlights the value of preserving clean benchmark snapshots and calibrated historical fits for ongoing leaderboard auditing.


\section{Discussion}
\label{sec:discussion}

\textsc{DualEval} suggests a shift from treating evaluation items as interchangeable test cases to treating them as calibrated measurement instruments. Once items have learned difficulty, sharpness, informativeness, and residual profiles, benchmark maintenance becomes a model-item problem: evaluators can identify saturated items, preserve high-signal subsets, and audit suspicious outcomes rather than relying only on aggregate leaderboard scores.

This perspective also points to richer extensions. A scalar ability parameter is useful for interpretability and ranking, but many evaluation domains contain multiple skills; multidimensional IRT provides a natural path toward modeling specialization across reasoning, coding, factuality, and interaction quality. More broadly, calibrated item profiles could support adaptive evaluation protocols that choose questions based on the current uncertainty about a model, while retaining item-level evidence for why a ranking changed.

\section{Conclusion}
We introduced \textsc{DualEval}, a framework that jointly models static benchmark correctness and arena-style preference signals on a shared latent scale. Across four domains, \textsc{DualEval} produces interpretable model rankings while estimating item difficulty, sharpness, and informativeness. Baseline comparisons show that combining static and arena supervision yields a more balanced signal than either source alone, while item-level analyses support sample-efficient evaluation and diagnostic benchmark auditing. Together, these results point toward evaluation pipelines that jointly rank models and calibrate benchmark items, making leaderboards more efficient, interpretable, and grounded in item-level evidence.

\section*{Limitations}
Our study has several limitations. First, the primary arena supervision relies on our internal reward model, which limits direct reproducibility even though we describe its architecture, training objective, and training-data scale. We address this by validating the RM against held-out human preferences (Appendix~\S\ref{appendix:rm-verify}) and replicating the main analyses with the publicly available Skywork-Reward-V2-Qwen3-8B model (Appendix~\S\ref{appendix:cross-rm}), but broader validation across multiple public RMs and human preference annotations remains important.

Second, each model-question pair is evaluated with a single generated response. This matches many leaderboard settings but does not capture within-model sampling variability, sensitivity to decoding parameters, or multi-attempt performance. Extending \textsc{DualEval} to repeated responses would allow item sharpness and residual anomalies to separate model capability from response-level stochasticity.

Finally, the current formulation uses a scalar ability parameter within each domain. This keeps rankings interpretable, but future evaluation systems should move toward multidimensional model profiles that capture specialization across skills, task formats, and interaction settings. Such extensions could make item selection adaptive not only to overall model strength, but also to the specific capabilities an evaluator wants to measure.


\bibliography{main}

@inproceedings{chiang2024chatbot,
  title={Chatbot arena: An open platform for evaluating llms by human preference},
  author={Chiang, Wei-Lin and Zheng, Lianmin and Sheng, Ying and Angelopoulos, Anastasios Nikolas and Li, Tianle and Li, Dacheng and Zhu, Banghua and Zhang, Hao and Jordan, Michael and Gonzalez, Joseph E and others},
  booktitle={Forty-first International Conference on Machine Learning},
  year={2024}
}

@article{magar2022data,
  title={Data contamination: From memorization to exploitation},
  author={Magar, Inbal and Schwartz, Roy},
  journal={arXiv preprint arXiv:2203.08242},
  year={2022}
}

@article{ott2022mapping,
  title={Mapping global dynamics of benchmark creation and saturation in artificial intelligence},
  author={Ott, Simon and Barbosa-Silva, Adriano and Blagec, Kathrin and Brauner, Jan and Samwald, Matthias},
  journal={Nature Communications},
  volume={13},
  number={1},
  pages={6793},
  year={2022},
  publisher={Nature Publishing Group UK London}
}

@article{li2024crowdsourced,
  title={From crowdsourced data to high-quality benchmarks: Arena-hard and benchbuilder pipeline},
  author={Li, Tianle and Chiang, Wei-Lin and Frick, Evan and Dunlap, Lisa and Wu, Tianhao and Zhu, Banghua and Gonzalez, Joseph E and Stoica, Ion},
  journal={arXiv preprint arXiv:2406.11939},
  year={2024}
}

@article{bradley1952rank,
  title={Rank analysis of incomplete block designs: I. the method of paired comparisons},
  author={Bradley, Ralph Allan and Terry, Milton E},
  journal={Biometrika},
  volume={39},
  number={3/4},
  pages={324--345},
  year={1952},
  publisher={JSTOR}
}

@book{lord2008statistical,
  title={Statistical theories of mental test scores},
  author={Lord, Frederic M and Novick, Melvin R},
  year={2008},
  publisher={IAP}
}

@article{ackerman2003using,
  title={Using multidimensional item response theory to evaluate educational and psychological tests},
  author={Ackerman, Terry A and Gierl, Mark J and Walker, Cindy M},
  journal={Educational Measurement: Issues and Practice},
  volume={22},
  number={3},
  pages={37--51},
  year={2003},
  publisher={Wiley Online Library}
}

@article{hendrycks2020measuring,
  title={Measuring massive multitask language understanding},
  author={Hendrycks, Dan and Burns, Collin and Basart, Steven and Zou, Andy and Mazeika, Mantas and Song, Dawn and Steinhardt, Jacob},
  journal={arXiv preprint arXiv:2009.03300},
  year={2020}
}

@article{hendrycks2021measuring,
  title={Measuring mathematical problem solving with the math dataset},
  author={Hendrycks, Dan and Burns, Collin and Kadavath, Saurav and Arora, Akul and Basart, Steven and Tang, Eric and Song, Dawn and Steinhardt, Jacob},
  journal={arXiv preprint arXiv:2103.03874},
  year={2021}
}

@article{kwiatkowski2019natural,
  title={Natural questions: a benchmark for question answering research},
  author={Kwiatkowski, Tom and Palomaki, Jennimaria and Redfield, Olivia and Collins, Michael and Parikh, Ankur and Alberti, Chris and Epstein, Danielle and Polosukhin, Illia and Devlin, Jacob and Lee, Kenton and others},
  journal={Transactions of the Association for Computational Linguistics},
  volume={7},
  pages={453--466},
  year={2019},
  publisher={MIT Press One Rogers Street, Cambridge, MA 02142-1209, USA journals-info~…}
}

@article{rajpurkar2016squad,
  title={Squad: 100,000+ questions for machine comprehension of text},
  author={Rajpurkar, Pranav and Zhang, Jian and Lopyrev, Konstantin and Liang, Percy},
  journal={arXiv preprint arXiv:1606.05250},
  year={2016}
}

@inproceedings{yang2018hotpotqa,
  title={HotpotQA: A dataset for diverse, explainable multi-hop question answering},
  author={Yang, Zhilin and Qi, Peng and Zhang, Saizheng and Bengio, Yoshua and Cohen, William and Salakhutdinov, Ruslan and Manning, Christopher D},
  booktitle={Proceedings of the 2018 conference on empirical methods in natural language processing},
  pages={2369--2380},
  year={2018}
}

@article{cobbe2021training,
  title={Training verifiers to solve math word problems},
  author={Cobbe, Karl and Kosaraju, Vineet and Bavarian, Mohammad and Chen, Mark and Jun, Heewoo and Kaiser, Lukasz and Plappert, Matthias and Tworek, Jerry and Hilton, Jacob and Nakano, Reiichiro and others},
  journal={arXiv preprint arXiv:2110.14168},
  year={2021}
}

@inproceedings{bowman2015large,
  title={A large annotated corpus for learning natural language inference},
  author={Bowman, Samuel and Angeli, Gabor and Potts, Christopher and Manning, Christopher D},
  booktitle={Proceedings of the 2015 conference on empirical methods in natural language processing},
  pages={632--642},
  year={2015}
}

@inproceedings{williams2018broad,
  title={A broad-coverage challenge corpus for sentence understanding through inference},
  author={Williams, Adina and Nangia, Nikita and Bowman, Samuel},
  booktitle={Proceedings of the 2018 conference of the North American chapter of the association for computational linguistics: human language technologies, volume 1 (long papers)},
  pages={1112--1122},
  year={2018}
}

@inproceedings{wang2018glue,
  title={GLUE: A multi-task benchmark and analysis platform for natural language understanding},
  author={Wang, Alex and Singh, Amanpreet and Michael, Julian and Hill, Felix and Levy, Omer and Bowman, Samuel},
  booktitle={Proceedings of the 2018 EMNLP workshop BlackboxNLP: Analyzing and interpreting neural networks for NLP},
  pages={353--355},
  year={2018}
}

@article{zheng2023judging,
  title={Judging llm-as-a-judge with mt-bench and chatbot arena},
  author={Zheng, Lianmin and Chiang, Wei-Lin and Sheng, Ying and Zhuang, Siyuan and Wu, Zhanghao and Zhuang, Yonghao and Lin, Zi and Li, Zhuohan and Li, Dacheng and Xing, Eric and others},
  journal={Advances in neural information processing systems},
  volume={36},
  pages={46595--46623},
  year={2023}
}

@article{liang2022holistic,
  title={Holistic evaluation of language models},
  author={Liang, Percy and Bommasani, Rishi and Lee, Tony and Tsipras, Dimitris and Soylu, Dilara and Yasunaga, Michihiro and Zhang, Yian and Narayanan, Deepak and Wu, Yuhuai and Kumar, Ananya and others},
  journal={arXiv preprint arXiv:2211.09110},
  year={2022}
}

@book{embretson2013item,
  title={Item response theory for psychologists},
  author={Embretson, Susan E and Reise, Steven P},
  year={2013},
  publisher={Psychology Press}
}

@inproceedings{lalor2019learning,
  title={Learning latent parameters without human response patterns: Item response theory with artificial crowds},
  author={Lalor, John P and Wu, Hao and Yu, Hong},
  booktitle={Proceedings of the Conference on Empirical Methods in Natural Language Processing. Conference on Empirical Methods in Natural Language Processing},
  volume={2019},
  pages={4240},
  year={2019}
}

@article{martinez2019item,
  title={Item response theory in AI: Analysing machine learning classifiers at the instance level},
  author={Mart{\'\i}nez-Plumed, Fernando and Prud{\^e}ncio, Ricardo BC and Mart{\'\i}nez-Us{\'o}, Adolfo and Hern{\'a}ndez-Orallo, Jos{\'e}},
  journal={Artificial intelligence},
  volume={271},
  pages={18--42},
  year={2019},
  publisher={Elsevier}
}

@article{liao2025toward,
  title={Toward a unified framework for data-efficient evaluation of large language models},
  author={Liao, Lele and Zhang, Qile and Wu, Ruofan and Fang, Guanhua},
  journal={arXiv preprint arXiv:2510.04051},
  year={2025}
}

@article{ni2024mixeval,
  title={Mixeval: Deriving wisdom of the crowd from llm benchmark mixtures},
  author={Ni, Jinjie and Xue, Fuzhao and Yue, Xiang and Deng, Yuntian and Shah, Mahir and Jain, Kabir and Neubig, Graham and You, Yang},
  journal={Advances in Neural Information Processing Systems},
  volume={37},
  pages={98180--98212},
  year={2024}
}

@article{robertson2025identity,
  title={Identity-Link IRT for Label-Free LLM Evaluation: Preserving Additivity in TVD-MI Scores},
  author={Robertson, Zachary},
  journal={arXiv preprint arXiv:2510.14966},
  year={2025}
}

@article{chen2021evaluating,
  title={Evaluating large language models trained on code},
  author={Chen, Mark and Tworek, Jerry and Jun, Heewoo and Yuan, Qiming and Pinto, Henrique Ponde De Oliveira and Kaplan, Jared and Edwards, Harri and Burda, Yuri and Joseph, Nicholas and Brockman, Greg and others},
  journal={arXiv preprint arXiv:2107.03374},
  year={2021}
}

@article{austin2021program,
  title={Program synthesis with large language models},
  author={Austin, Jacob and Odena, Augustus and Nye, Maxwell and Bosma, Maarten and Michalewski, Henryk and Dohan, David and Jiang, Ellen and Cai, Carrie and Terry, Michael and Le, Quoc and others},
  journal={arXiv preprint arXiv:2108.07732},
  year={2021}
}

@inproceedings{zhou2026lost,
  title={Lost in benchmarks? rethinking large language model benchmarking with item response theory},
  author={Zhou, Hongli and Huang, Hui and Zhao, Ziqing and Han, Lvyuan and Wang, Huicheng and Chen, Kehai and Yang, Muyun and Bao, Wei and Dong, Jian and Xu, Bing and others},
  booktitle={Proceedings of the AAAI Conference on Artificial Intelligence},
  volume={40},
  pages={35085--35093},
  year={2026}
}

@article{li2025adaptive,
  title={Adaptive Testing for LLM Evaluation: A Psychometric Alternative to Static Benchmarks},
  author={Li, Peiyu and Tang, Xiuxiu and Chen, Si and Cheng, Ying and Metoyer, Ronald and Hua, Ting and Chawla, Nitesh V},
  journal={arXiv preprint arXiv:2511.04689},
  year={2025}
}

@inproceedings{wang2026metaeval,
  title={MetaEval: Measuring the Discrimination of Benchmarks for Efficient LLM Evaluation},
  author={Wang, Zhuo and Wu, Wen and Wang, Guoqing and Ye, Guangze and Cheng, Zhenxiao},
  booktitle={Proceedings of the AAAI Conference on Artificial Intelligence},
  volume={40},
  pages={33773--33781},
  year={2026}
}

@article{wang2025rethinking,
  title={Rethinking LLM Evaluation: Can We Evaluate LLMs with 200x Less Data?},
  author={Wang, Shaobo and Wang, Cong and Fu, Wenjie and Min, Yue and Feng, Mingquan and Guan, Isabel and Hu, Xuming and He, Conghui and Wang, Cunxiang and Yang, Kexin and others},
  journal={arXiv preprint arXiv:2510.10457},
  year={2025}
}

@article{castleman2025rethinking,
  title={Rethinking math benchmarks for llms using irt},
  author={Castleman, Jane and Nadeem, Nimra and Namjoshi, Tanvi and Liu, Lydia T},
  journal={Proceedings of Machine Learning Research},
  volume={273},
  pages={66--82},
  year={2025},
  publisher={ML Research Press}
}

@article{merrill2026terminal,
  title={Terminal-bench: Benchmarking agents on hard, realistic tasks in command line interfaces},
  author={Merrill, Mike A and Shaw, Alexander G and Carlini, Nicholas and Li, Boxuan and Raj, Harsh and Bercovich, Ivan and Shi, Lin and Shin, Jeong Yeon and Walshe, Thomas and Buchanan, E Kelly and others},
  journal={arXiv preprint arXiv:2601.11868},
  year={2026}
}

@inproceedings{jimenez2024swe,
  title={Swe-bench: Can language models resolve real-world github issues?},
  author={Jimenez, Carlos E and Yang, John and Wettig, Alexander and Yao, Shunyu and Pei, Kexin and Press, Ofir and Narasimhan, Karthik},
  booktitle={International Conference on Learning Representations},
  volume={2024},
  pages={54107--54157},
  year={2024}
}

@inproceedings{van2019best,
  title={Best practices for the human evaluation of automatically generated text},
  author={Van Der Lee, Chris and Gatt, Albert and Van Miltenburg, Emiel and Wubben, Sander and Krahmer, Emiel},
  booktitle={Proceedings of the 12th international conference on natural language generation},
  pages={355--368},
  year={2019}
}

@inproceedings{basile2021we,
  title={We need to consider disagreement in evaluation},
  author={Basile, Valerio and Fell, Michael and Fornaciari, Tommaso and Hovy, Dirk and Paun, Silviu and Plank, Barbara and Poesio, Massimo and Uma, Alexandra},
  booktitle={Proceedings of the 1st workshop on benchmarking: past, present and future},
  pages={15--21},
  year={2021}
}

@inproceedings{jain2025livecodebench,
  title={Livecodebench: Holistic and contamination free evaluation of large language models for code},
  author={Jain, Naman and Gu, Alex and Li, Wen-Ding and Yan, Fanjia and Zhang, Tianjun and Wang, Sida and Solar-Lezama, Armando and Sen, Koushik and Stoica, Ion},
  booktitle={International Conference on Learning Representations},
  volume={2025},
  pages={58791--58831},
  year={2025}
}

@article{phan2025humanity,
  title={Humanity's last exam},
  author={Phan, Long and Gatti, Alice and Han, Ziwen and Li, Nathaniel and Hu, Josephina and Zhang, Hugh and Zhang, Chen Bo Calvin and Shaaban, Mohamed and Ling, John and Shi, Sean and others},
  journal={arXiv preprint arXiv:2501.14249},
  year={2025}
}

@inproceedings{he2024olympiadbench,
  title={Olympiadbench: A challenging benchmark for promoting agi with olympiad-level bilingual multimodal scientific problems},
  author={He, Chaoqun and Luo, Renjie and Bai, Yuzhuo and Hu, Shengding and Thai, Zhen and Shen, Junhao and Hu, Jinyi and Han, Xu and Huang, Yujie and Zhang, Yuxiang and others},
  booktitle={Proceedings of the 62nd Annual Meeting of the Association for Computational Linguistics (Volume 1: Long Papers)},
  pages={3828--3850},
  year={2024}
}

@article{wei2024measuring,
  title={Measuring short-form factuality in large language models},
  author={Wei, Jason and Karina, Nguyen and Chung, Hyung Won and Jiao, Yunxin Joy and Papay, Spencer and Glaese, Amelia and Schulman, John and Fedus, William},
  journal={arXiv preprint arXiv:2411.04368},
  year={2024}
}

@article{liu2023your,
  title={Is your code generated by chatgpt really correct? rigorous evaluation of large language models for code generation},
  author={Liu, Jiawei and Xia, Chunqiu Steven and Wang, Yuyao and Zhang, Lingming},
  journal={Advances in neural information processing systems},
  volume={36},
  pages={21558--21572},
  year={2023}
}

@misc{maa_aime,
  title={American Invitational Mathematics Examination (AIME)},
  author={{Mathematical Association of America}},
  year={2026},
  howpublished={MAA Invitational Competitions}
}

\appendix
\section{Additional \textsc{DualEval} Ranking Visualizations}
\label{appendix:more-rankings}

\begin{figure*}[t]
\centering
\includegraphics[width=0.92\linewidth,trim={0 0.2cm 0 0cm},clip]{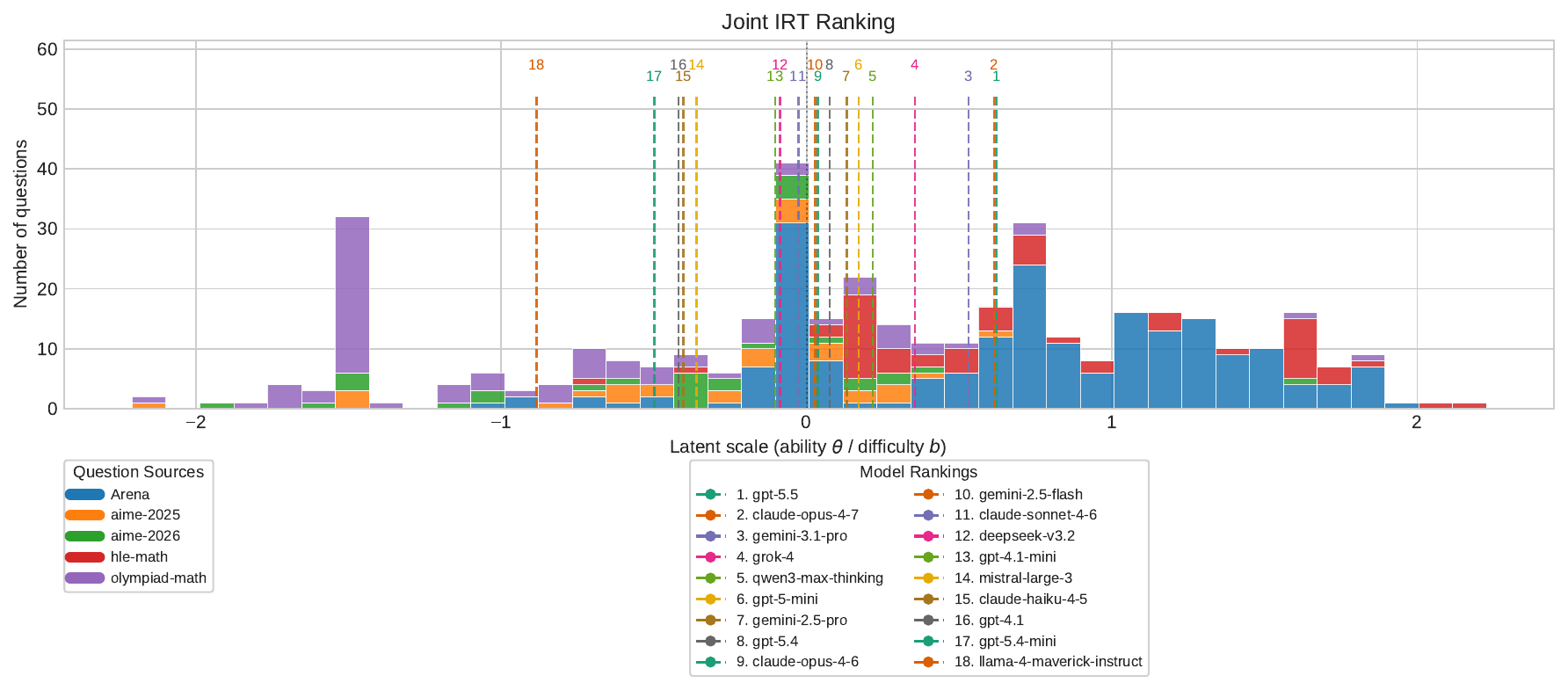}
\caption{
Learned model abilities and question difficulties on \textbf{math}.
}
\label{fig:math_rankings}
\end{figure*}

\begin{figure*}[t]
\centering
\includegraphics[width=0.92\linewidth,trim={0 0.2cm 0 0},clip]{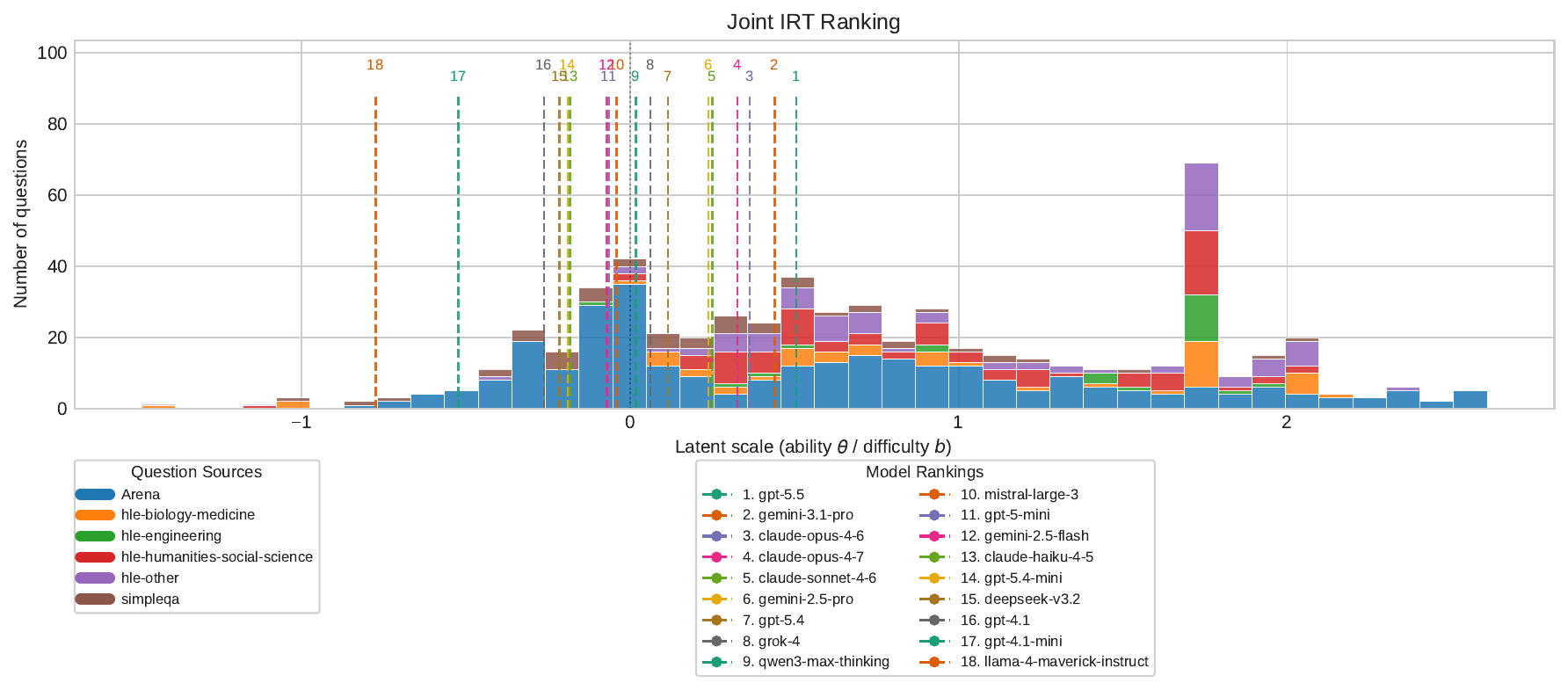}
\caption{
Learned model abilities and question difficulties on \textbf{miscellaneous}.
}
\label{fig:misc_rankings}
\vspace{-0.8em}
\end{figure*}

\begin{figure*}[t]
\centering
\includegraphics[width=0.92\linewidth,trim={0 0.2cm 0 0cm},clip]{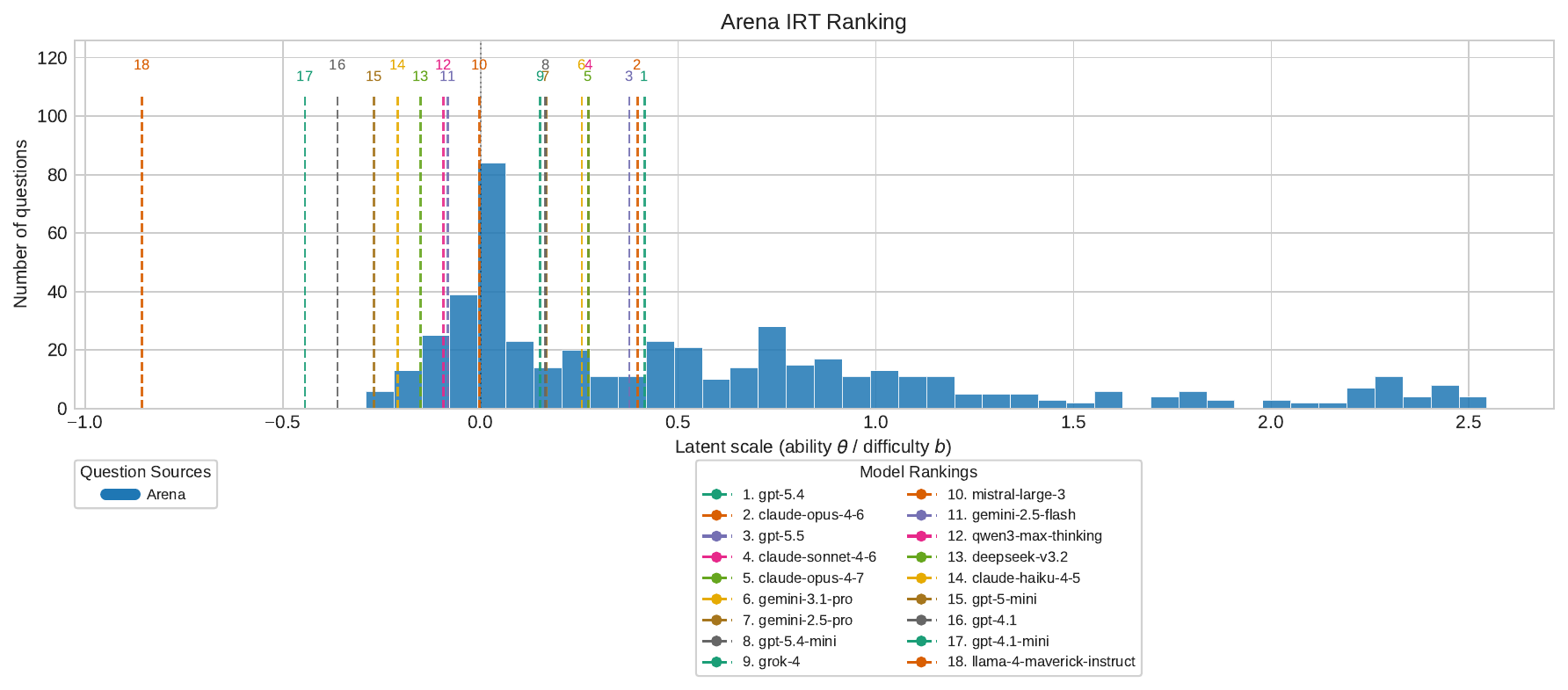}
\caption{
Learned model abilities and question difficulties on \textbf{generic}.
}
\label{fig:generic_rankings}
\end{figure*}

We provide additional examples of \textsc{DualEval} rankings of math, miscellaneous, and generic domains in Figures ~\ref{fig:math_rankings}, \ref{fig:misc_rankings}, and \ref{fig:generic_rankings}.

\begin{figure*}
    \centering
    \includegraphics[width=\linewidth]{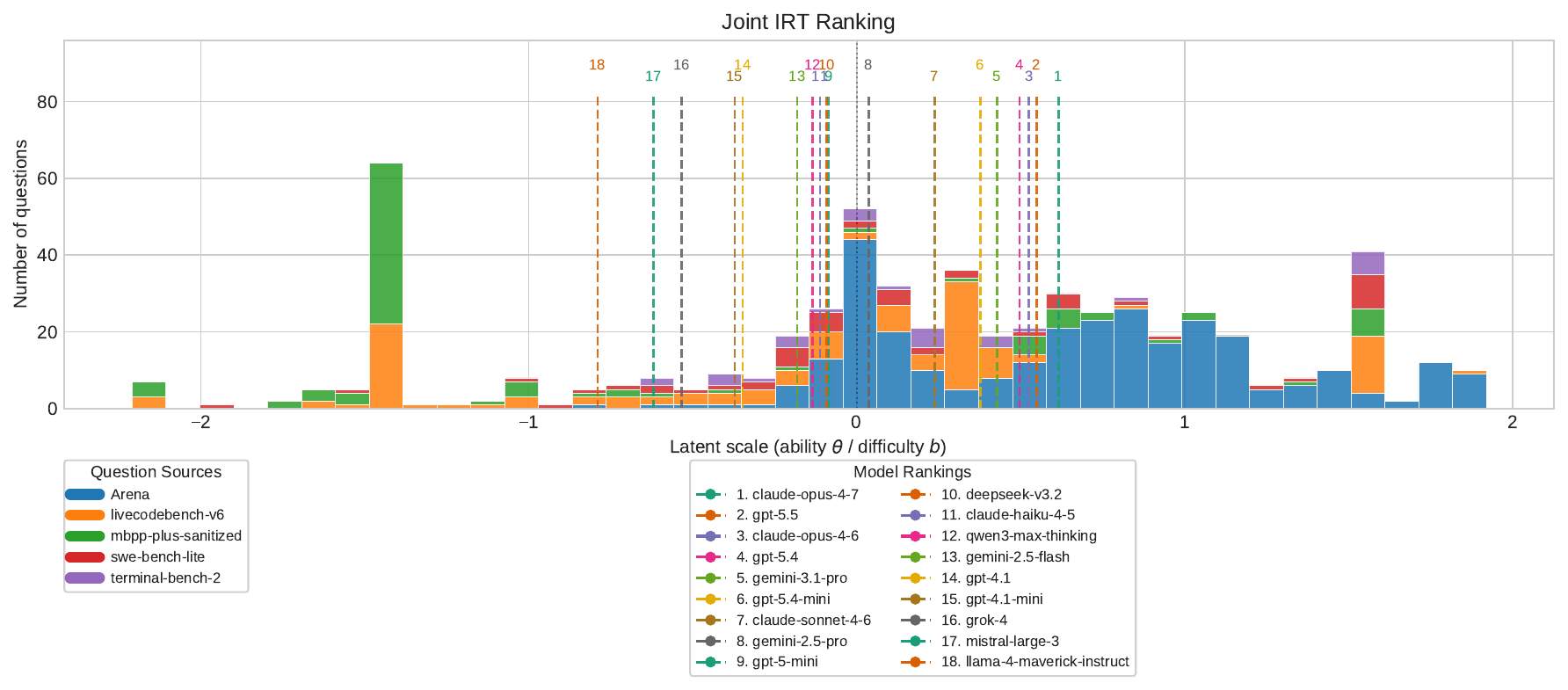}
    \caption{Rankings of model abilities and model-centric question difficulties on the \textbf{coding} domain, using public reward model \texttt{Skywork-Reward-V2-Qwen3-8B}.}
    \label{fig:coding_rankings_public}
\end{figure*}

\section{\textsc{DualEval} Hyperparameters and Configurations}
\label{appendix:hyperparams}

All \textsc{DualEval} experiments in this paper use the same hyperparameters unless otherwise noted: Adam optimizer with learning rate $0.02$, $2000$ epochs, loss weights $\lambda_{\text{static}}=\lambda_{\text{pref}}=1.0$ and $\lambda_{\text{bb}}=0.2$, L2 regularization coefficient $\lambda_{\text{reg}}=0.01$ applied to $\theta$, $b$, $k$, and $\log \tilde\gamma$, and pair-treatment thresholds derived as the 15\textsuperscript{th} percentile of $\max(z_i, z_j)$ for both-bad flagging and the 15\textsuperscript{th} percentile of $|z_i - z_j|$ for tie filtering, applied per-domain.  The arena temperature is parameterized as $\gamma = \gamma_0 \cdot \tilde\gamma$ with $\gamma_0 = 4$, and $\log \tilde\gamma$ is L2-regularized toward zero.  After each optimizer step, model abilities are zero-mean centered with the same shift absorbed into question difficulties, preserving all differences $\theta_i - b_q$. All fits use single-precision floating point on CPU; a single domain's joint fit completes in 10--25 seconds. Arena-only fits on the generic domain use the same configuration with the static loss disabled.

\section{Validation of Reward Model with Real Human Preference}
\label{appendix:rm-verify}

This appendix supports the robustness claims in Section~\ref{sec:experiments} by (i) quantifying how well each reward model agrees with real human preference on the
prompts actually used by our pipeline, (ii) visualizing cross-RM agreement of \textsc{DualEval}'s learned model abilities in the static-anchored domains, and (iii)
documenting the arena-only generic domain as a no-anchor reference.

\paragraph{Reward-model alignment with human preference.}
For each domain we identify a subset of arena prompts for which human preference labels are available, and that have a decisive (non-tie) human winner. Sample sizes per domain are reported in Table~\ref{tab:rm-human-alignment}. Both reward models are scored on the same pairs from this filtered subset. Table~\ref{tab:rm-human-alignment}
reports the per-domain pair count and the RM's accuracy at predicting the human winner. Our proprietary RM is substantially more human-aligned than the public RM in
every domain.

\begin{table}[ht]
\centering
\small
\setlength{\tabcolsep}{6pt}
\begin{tabular}{l r c c}
\toprule
\textbf{Domain} & \textbf{N} & \textbf{Proprietary RM} & \textbf{Public RM} \\
\midrule  
Coding  & 195 & 0.739 & 0.573 \\
Math    & 120 & 0.875 & 0.708 \\
Misc    & 200 & 0.715 & 0.655 \\
Generic & 376 & 0.742 & 0.596 \\
\bottomrule
\end{tabular}
\caption{Pairwise accuracy of each reward model at predicting the human winner. Restricted to pairs with a decisive (non-tie) human-preference winner.}
\label{tab:rm-human-alignment}
\end{table}



\section{Stability of DualEval Rankings}
\label{appendix:stability}

We assess the stability of \textsc{DualEval} rankings using a question-level cluster bootstrap.  For each domain we draw $B=100$ bootstrap replicates, resampling questions with replacement stratified by source (each benchmark source plus arena), so every replicate has the same per-source $N$ as the original.  Drawn copies of a question receive unique IDs so the IRT fit treats them as independent items--the standard cluster-bootstrap behavior.  Reward $z$-scores are renormalized per replicate. For each replicate we refit \textsc{DualEval} with the published hyperparameters and record per-model $\hat\theta_i$.  We use the same procedure to bootstrap Static 2PL and Arena BT for direct comparison (Static 2PL is omitted for the arena-only generic domain). 

\begin{figure*}[t]
\centering
\includegraphics[width=\linewidth]{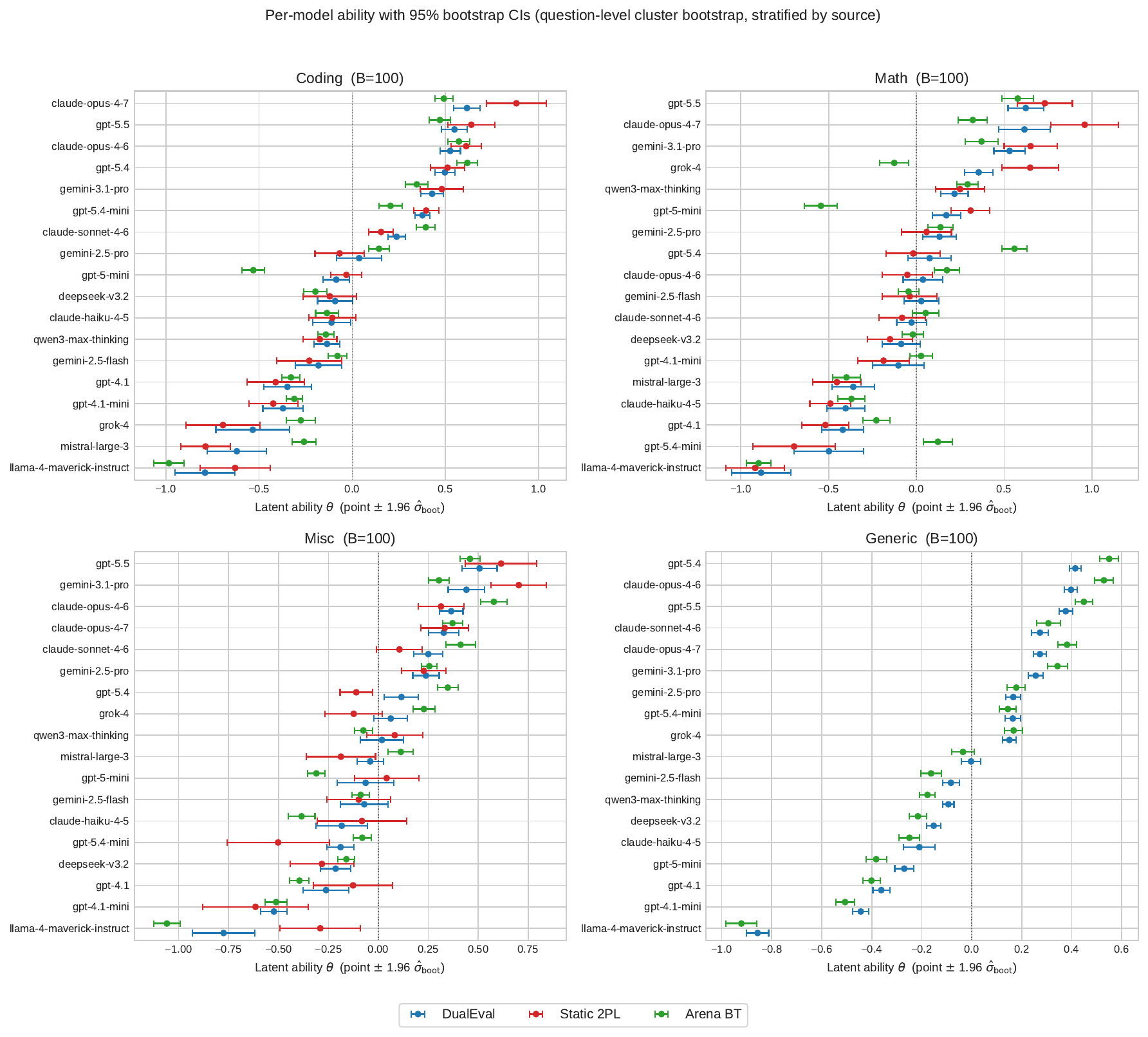}
\caption{Per-model ability with 95\% bootstrap CIs (question-level cluster bootstrap, $B=100$, stratified by source).}
\label{fig:stability}
\end{figure*}

Figure~\ref{fig:stability} shows per-model $\hat\theta_i$ with symmetric 95\% CIs ($\hat\theta_i \pm 1.96,\hat\sigma_{\text{boot}}$) for all three methods in each domain.  Top-3 and bottom-3 positions are stable to $\pm 1$ rank across replicates in every domain.  Middle-band positions span 4--6 ranks of 95\% CI in the three static-anchored domains, indicating that not all adjacent leaderboard positions are statistically distinguishable at the current evaluation size.  Generic--being arena-only with 500 prompts contributing $\binom{18}{2}$ pairwise comparisons per item--produces the tightest CIs across the four domains.

Table~\ref{tab:bootstrap_method_compare} reports per-(domain, method) summary statistics.  On equivalent data (generic, arena-only), \textsc{DualEval} matches or slightly improves over Arena BT (median $\hat\sigma_\theta = 0.016$ vs $0.019$), confirming that the IRT structure adds discriminative information without inflating uncertainty.  Against Static 2PL on the static-anchored domains, the joint formulation reduces ranking variance by 30--60\%. Against Arena BT on those domains, \textsc{DualEval} carries a modest variance premium that reflects the additional cross-source information being integrated, in exchange for substantially better Static-$\rho$ (Table \ref{tab:integration}).

\begin{table}[t]
\centering
\small
\setlength{\tabcolsep}{5pt}
\begin{tabular}{l l c c}
\toprule
Domain & Method & Median $\hat\sigma_\theta$ & Median rank-CI \\
\midrule
Coding & Static 2PL & 0.065 & 3.0 \\
& Arena BT & 0.028 & 1.0 \\
& DualEval & 0.040 & 2.0 \\
\midrule
Math & Static 2PL & 0.074 & 3.0 \\
& Arena BT & 0.039 & 2.0 \\
& DualEval & 0.055 & 3.0 \\
\midrule
Misc & Static 2PL & 0.082 & 4.5 \\
& Arena BT & 0.026 & 1.0 \\
& DualEval & 0.043 & 3.0 \\
\midrule
Generic & Arena BT & 0.019 & 1.0 \\
& DualEval & 0.016 & 1.0 \\
\bottomrule
\end{tabular}
\caption{Bootstrap uncertainty per (domain, method) under a question-level cluster bootstrap
(B=100, stratified by source). $\hat\sigma_\theta$ is the bootstrap standard deviation of
model ability; median rank-CI is the median 95\% bootstrap rank interval width over the 18
models. Static 2PL is omitted for Generic (arena-only). On equivalent data (Generic),
DualEval matches or improves over Arena BT; against Static 2PL the joint formulation reduces
ranking variance by 30--60\%.}
\label{tab:bootstrap_method_compare}
\end{table}

\section{Held-out calibration}
\label{appendix:calibration}

\begin{figure*}[t]
\centering
\includegraphics[width=\linewidth]{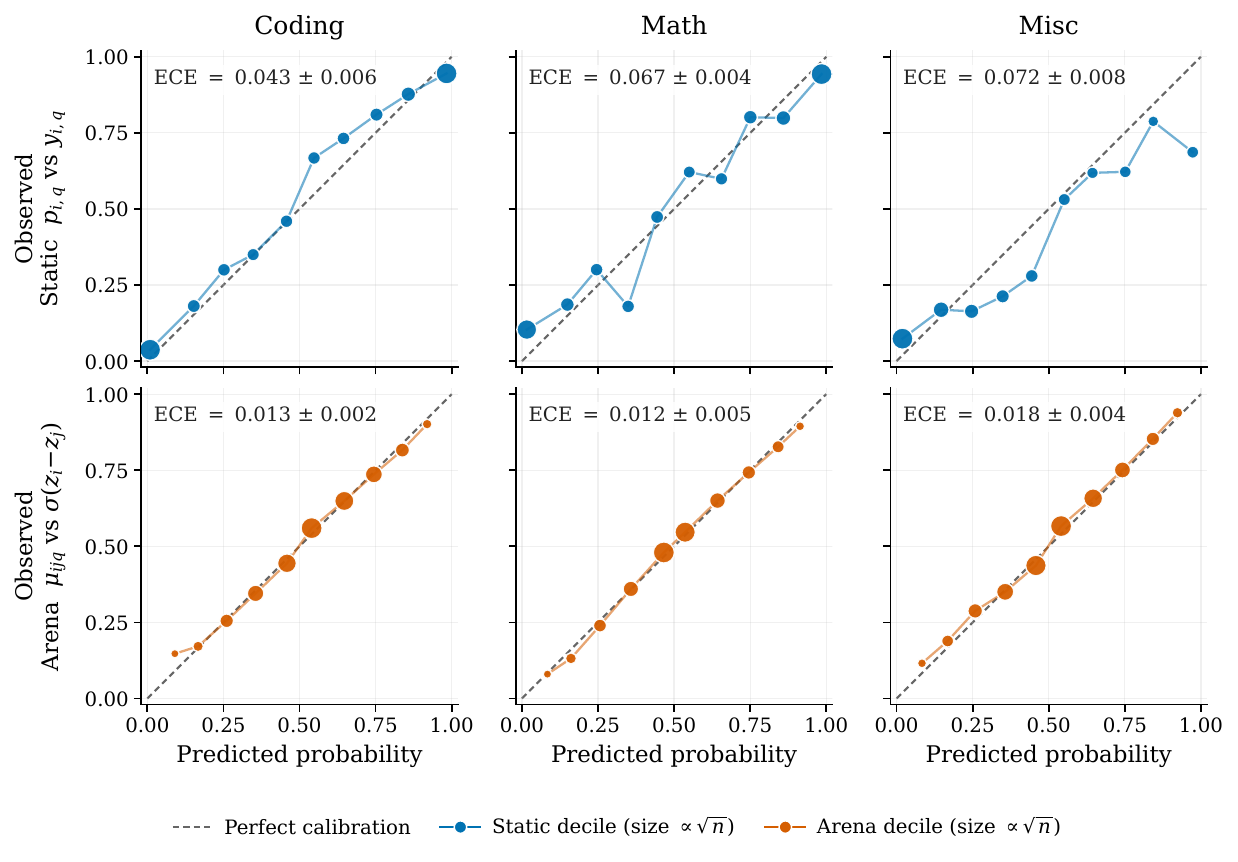}
\caption{Held-out calibration of DualEval predictions. Rows: static cells ($p_{i,q}$ against
binary correctness) and arena pairs ($\mu_{ijq}$ against soft pairwise targets
$\sigma(z_i-z_j)$). Columns: coding, math, miscellaneous domains. Each point is a decile of
the predicted-probability axis, with size $\propto \sqrt{n}$. Dashed line marks perfect
calibration. ECE (mean $\pm$ SD over three seeds) annotated per panel.}
\label{fig:calibration}
\end{figure*}

We evaluate calibration of the joint fit's per-pair predictions on a cell-level 80/20 random holdout (averaged over three random seeds; same fit hyperparameters as \S\ref{sec:experiments}). On held-out static cells, predicted success probabilities $p_{i,q}$ match observed binary outcomes with expected calibration error (ECE) of $0.04$--$0.07$ across coding/math/misc domains. On held-out arena pairs, predicted preference probabilities $\mu_{ijq}$ match the soft pairwise targets $\sigma(z_{i,q}-z_{j,q})$ with ECE of $0.01$--$0.03$, with reliability diagrams nearly diagonal across the predicted-probability range (Figure~\ref{fig:calibration}). For transparency we also report ECE against the binarised pairwise outcome $\mathbf{1}[\,\sigma(z_{i,q}-z_{j,q}) \geq 0.5\,]$, which is higher ($0.07$--$0.16$) due to expected under-confidence when soft pairwise targets near $0.5$ are discretised to hard win/loss labels; this binarisation does not reflect a miscalibration of the underlying joint fit. These calibration results justify treating $p_{i,q}$ and $\mu_{ijq}$ as meaningful probability estimates in the downstream Fisher information (\S\ref{sec:item-informativeness}) and residual diagnostics
(\S\ref{sec:anomaly-detection}).

\section{Cross-RM Ranking Agreement}
\label{appendix:cross-rm}
\begin{figure*}[ht]
\centering
\includegraphics[width=\linewidth]{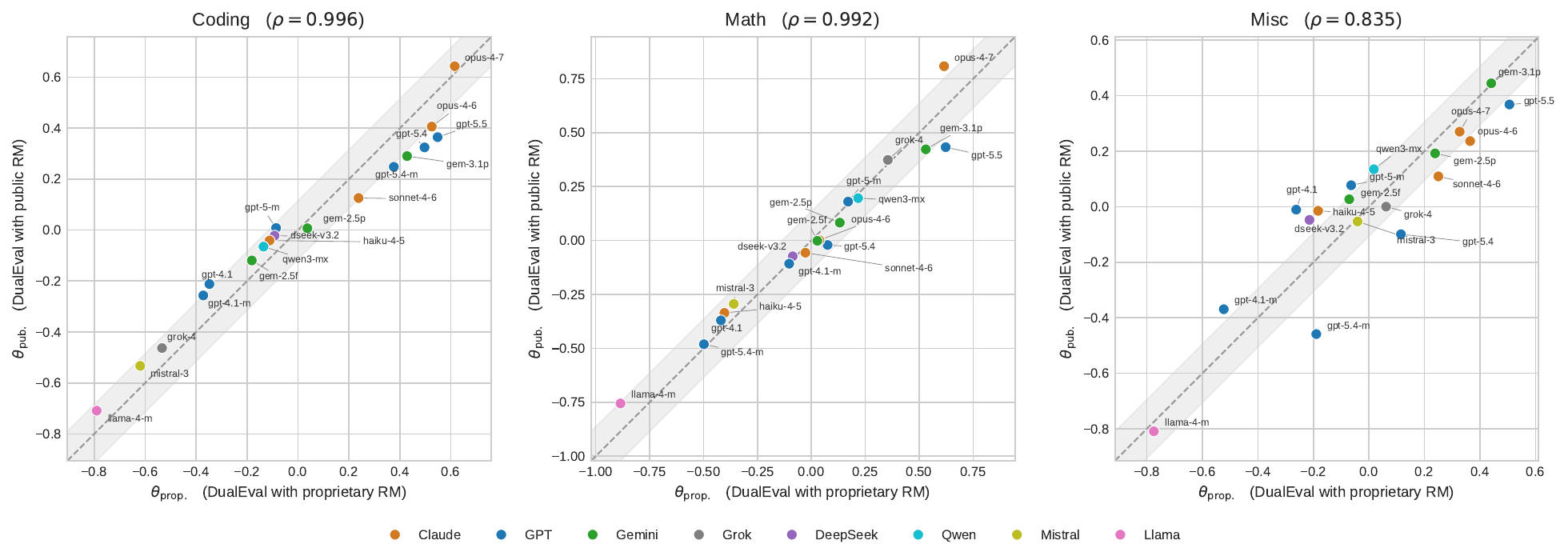}
\caption{Per-domain scatter of \textsc{DualEval} model abilities under the proprietary RM ($x$-axis) and the public RM ($y$-axis), three panels for coding/math/misc.
Points are individual frontier LLMs. Static anchoring keeps the cross-RM rankings tightly aligned in static-anchored domains.}
\label{fig:theta-scatter-rm}
\end{figure*}

Figure~\ref{fig:theta-scatter-rm} plots \textsc{DualEval}'s learned model abilities under the proprietary and public RMs against each other, one panel per static-anchored domain. Points fall close to the diagonal: Spearman correlations are $0.963$ (coding), $0.992$ (math), and $0.835$ (misc), consistent with the Static $\rho \geq 0.96$ entries in Table~\ref{tab:integration} and indicating that static anchoring stabilizes the ranking against RM-induced noise. The arena-only generic domain provides the no-anchor counterfactual: with no static labels to pin the latent scale, the agreement between the proprietary and public RMs drops to $\rho = 0.25$, empirically isolating static anchoring as the mechanism behind the joint framework's cross-RM stability.

\section{Evaluation Details}
\label{appendix:inference}

\paragraph{Benchmark sources.}
Our static benchmark pool draws from LiveCodeBench~\citep{jain2025livecodebench}, MBPP-Plus from EvalPlus~\citep{liu2023your}, SWE-Bench~\citep{jimenez2024swe}, TerminalBench~\citep{merrill2026terminal}, AIME 2025/2026 \citep{maa_aime}, Humanity's Last Exam (HLE)~\citep{phan2025humanity}, OlympiadBench/Olympiad-Math~\citep{he2024olympiadbench}, and SimpleQA~\citep{wei2024measuring}. Arena prompts are drawn from publicly available LMArena~\citep{chiang2024chatbot} logs and filtered by domain as described below.

\paragraph{Prompt sampling and filtering.}
Static items are drawn from each source benchmark by seeded stratified sampling, with strata defined by the natural axes of each source: difficulty tier for LiveCodeBench-v6, problem type for AIME and Olympiad-Math, HLE level for HLE-Math, and field for HLE-Misc (Bio./Med., Engineering, Humanities/Social Science, Other); SimpleQA is stratified by topic. MBPP-Plus, SWE-Bench-lite, and TerminalBench-2.0 are taken in full up to the per-source quotas in Table~\ref{tab:data_models}, after dropping items shorter than 150 characters. Arena prompts are sampled from LMArena releases: \texttt{arena-expert-5k} for coding, math, and miscellaneous, and\texttt{arena-human-preference-140k} for generic. We first apply a coarse domain pre-filter using the dataset-provided occupational and category tags, then run an LLM judge (\texttt{gpt-5.4-mini}) with a domain-specific rubric that returns an inclusion label and a confidence score: coding requires the answer to depend on runnable code, math requires the core task to be mathematical reasoning or calculation, miscellaneous requires expert domain knowledge outside code/math, and generic requires a non-technical everyday task with no code or math content. Candidates with confidence below the per-domain threshold ($0.5$ for coding/math/misc, $0.8$ for generic) are dropped, and the top-$N$ surviving prompts by judge confidence are retained to match the per-domain arena pool sizes in Table~\ref{tab:data_models} (coding $300$, math $200$, miscellaneous $300$, generic $500$).

\paragraph{Evaluation setup.} For each (model, question) pair we generate a single response with default sampling settings for each provider's API as of submission (temperature and top-$p$ are model-default; no system prompt is prepended for arena prompts; a standard ``provide a complete, correct solution'' instruction is appended for coding/math/misc). Model identifiers are reported in Table \ref{tab:data_models}. For agentic SWE-Bench tasks we route every model through \texttt{mini-SWE-agent-v2} with a single attempt and the harness's default rollout budget; for TerminalBench-2.0 we use \texttt{Terminus2} under identical configuration for all models.

\paragraph{Static grading.} Coding tasks are graded by their reference execution harness (test-suite pass for MBPP-Plus and LiveCodeBench; resolved-instance flag for SWE-Bench; harness pass for TerminalBench). Math and miscellaneous items use exact-answer matching after a normalization pass on the final-answer line; ambiguous cases are resolved by an LLM judge with chain-of-thought disabled.

\paragraph{Arena prompts.} For coding, math, and miscellaneous we filter LMArena prompts to the target domain via LLM judges using a fixed rubric. The generic domain uses LM Arena's category-mix without domain filtering. Per-domain arena pool sizes also appear in Table~\ref{tab:data_models}.

\paragraph{Reward scoring.} Each model response is scored by a single forward pass of the relevant reward model (proprietary or \texttt{Skywork-Reward-V2-Qwen3-8B} for the replication). Rewards are global-standardised per (domain, RM) before pairwise targets are constructed.

\section{Difficulty and Sharpness}
\label{appendix:difficulty_sharpness}

\begin{figure*}[t]
\centering
\includegraphics[width=\linewidth]{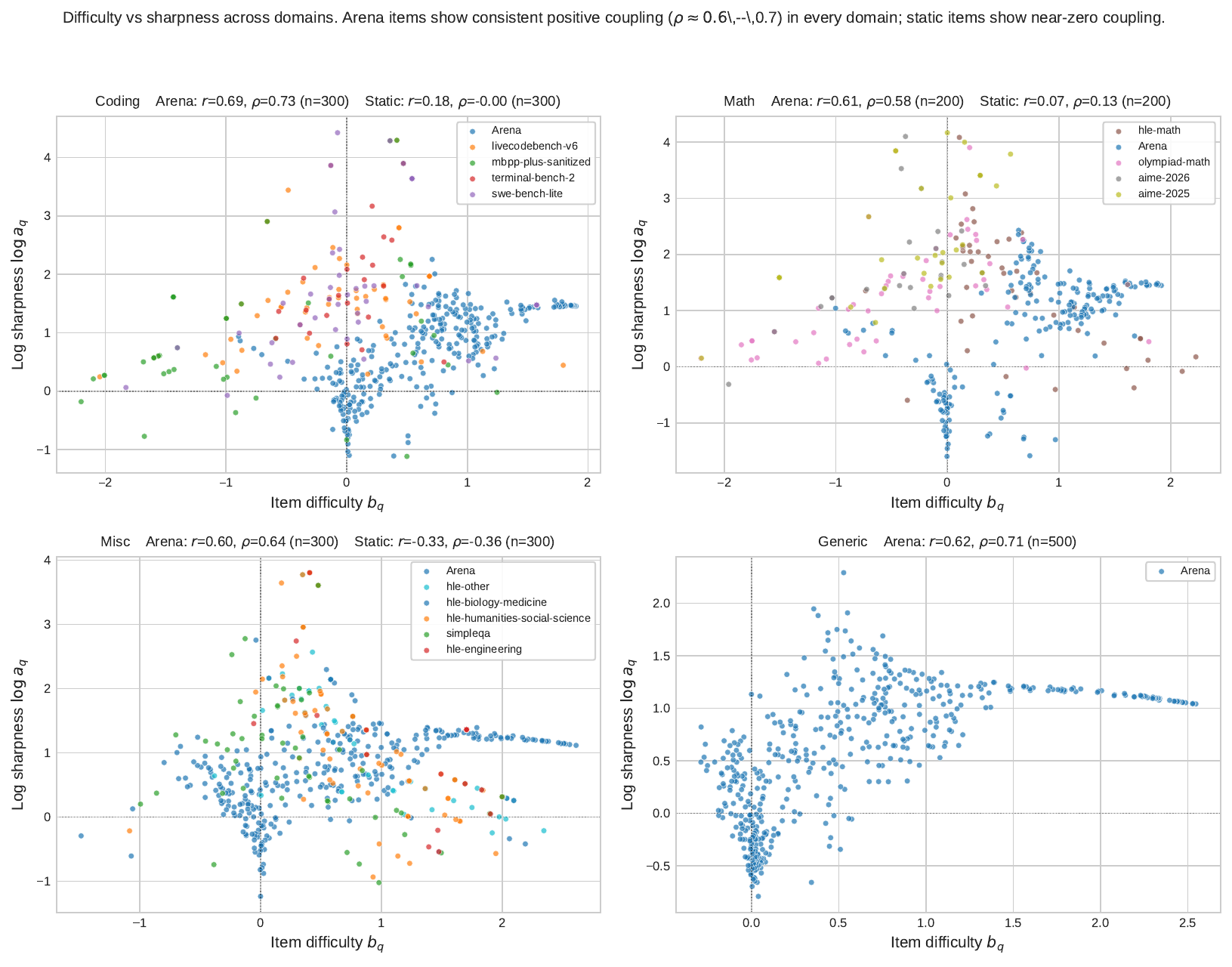}
\caption{Learned item difficulty versus sharpness across domains. Static items show weak difficulty--sharpness coupling, while arena items exhibit a stronger positive association, highlighting modality-specific item structure captured by \textsc{DualEval}.}
\label{fig:difficulty_sharpness}
\end{figure*}

Figure~\ref{fig:difficulty_sharpness} plots learned item difficulty $b_q$ against log-sharpness $\log a_q$ for every domain, with points colored by benchmark source.  Static items show near-zero or weakly negative correlation in every static-anchored domain ($\rho = -0.00, +0.13, -0.36$ for coding, math, misc), confirming that difficulty and sharpness are largely independent properties of a static item: saturated items can be sharp, hard items can be flat, and Fisher information cannot be inferred from difficulty alone. Arena items, by contrast, show a robust positive coupling ($\rho \in [+0.58, +0.73]$ across all four domains), reflecting how continuous reward signals propagate both location and spread into the IRT parameters. The joint \textsc{DualEval} formulation accommodates both regimes without imposing a single structural assumption on either, and the contrast itself confirms that the framework's learned parameters reflect the structure of the underlying outcome modality.

\section{Pair-Treatment Ablation}
\label{appendix:pair-treatment-ablation}

We ablate the two pair-treatment components introduced in \S\ref{sec:method}: both-bad (BB) anchoring and tie filtering. All variants are evaluated on the coding domain using the same 20\% held-out pair set constructed with the full configuration thresholds, so metrics are directly comparable. The full model uses BB flagging with $\lambda_{bb}=0.2$ and tie filtering.  Arena pair accuracy is measured on held-out decisive pairs; BB-AUC measures whether $1-\min(p_i,p_j)$ ranks BB pairs above non-BB pairs.

\begin{table*}[t]
\centering
\small
\setlength{\tabcolsep}{4pt}
\begin{tabular}{lcccc}
\toprule
\textbf{Variant} &
\textbf{BB Loss} &
\textbf{Tie Filter} &
\textbf{Arena Pair Acc.} &
\textbf{Both-Bad AUC} \\
\midrule
Full \textsc{DualEval} & \checkmark & \checkmark & \textbf{0.778} & \textbf{0.903} \\
No both-bad loss & -- & \checkmark & 0.771 & 0.622 \\
No both-bad filtering & -- & \checkmark & 0.774 & 0.627 \\
No tie filter & \checkmark & -- & 0.777 & 0.899 \\
No both-bad / no tie & -- & -- & 0.774 & 0.628 \\
\bottomrule
\end{tabular}
\caption{Coding-domain ablation of pair treatment for arena comparisons. Arena pair accuracy is measured on held-out hard pairs. Both-bad AUC measures whether the fitted model truly identifies pairs where both responses receive low rewards. }
\label{tab:bb_tie_ablation}
\end{table*}

Table~\ref{tab:bb_tie_ablation} shows that BB anchoring is the dominant choice: removing the BB loss drops BB-AUC from $0.903$ to $0.622$ while leaving arena pair accuracy near $0.78$, and removing BB flagging entirely gives essentially the same degradation. Tie filtering has a smaller, complementary effect --- it slightly improves BB calibration when combined with the full model (0.903 vs 0.899) but removing it alone barely changes either metric. The full configuration is used in all main-text experiments.

\section{Public Leaderboard Rankings}
\label{appendix:public_rankings}

Table~\ref{tab:public_rankings} lists the public LMArena Text-leaderboard ranks for the 18 models we evaluate, taken from the matched public Arena Text sub-leaderboards used in Table~2 (Coding, Math, Text-Expert, Text-Overall). These are the external reference rankings against which Table~\ref{tab:integration}'s \textit{Public $\rho$} column is computed. Lower rank is better. Public ranks were retrieved on the same date the paper's arena prompts were sampled.
  
\begin{table*}[t]
\centering
\small
\setlength{\tabcolsep}{5pt}
\begin{tabular}{lrrrr}
\toprule
Paper model & Text-Coding & Text-Math & Text-Expert & Text-Overall \\
\midrule
\texttt{claude-opus-4-6} & 4 & 5 & 2 & 3 \\
\texttt{claude-opus-4-7} & 2 & 9 & 3 & 4 \\
\texttt{gemini-3.1-pro} & 10 & 7 & 8 & 6 \\
\texttt{gpt-5.5} & 29 & 4 & 9 & 11 \\
\texttt{claude-sonnet-4-6} & 11 & 28 & 12 & 22 \\
\texttt{gpt-5.4} & 21 & 33 & 15 & 24 \\
\texttt{gpt-5.4-mini} & 38 & 53 & 27 & 39 \\
\texttt{gemini-2.5-pro} & 81 & 46 & 56 & 49 \\
\texttt{qwen3-max-thinking} & 60 & 52 & 50 & 62 \\
\texttt{deepseek-v3.2} & 76 & 63 & 71 & 76 \\
\texttt{mistral-large-3} & 75 & 110 & 105 & 93 \\
\texttt{gpt-4.1} & 97 & 146 & 118 & 94 \\
\texttt{gemini-2.5-flash} & 140 & 99 & 100 & 98 \\
\texttt{claude-haiku-4-5} & 64 & 125 & 69 & 99 \\
\texttt{grok-4} & 121 & 68 & 92 & 100 \\
\texttt{gpt-5-mini} & 134 & 101 & 122 & 124 \\
\texttt{gpt-4.1-mini} & 127 & 163 & 143 & 135 \\
\texttt{llama-4-maverick-instruct} & 186 & 180 & 181 & 197 \\
\bottomrule
\end{tabular}
\caption{
Public Arena ranks for the 18 models used in our external validation analysis.
Ranks are taken from the matched public Arena Text subleaderboards used in Table~\ref{tab:integration}:
Text-Coding, Text-Math, Text-Expert, and Text-Overall. Lower rank is better.
}
\label{tab:public_rankings}
\end{table*}

\section{Contamination-Detection Metric Details}
\label{appendix:contamination-metrics}

For each domain, modality, injection rate, and random seed, the synthetic contamination protocol defines a binary mask $m_{i,q}\in\{0,1\}$ over model--item cells, where $m_{i,q}=1$ denotes an injected cell. Each detector assigns a scalar anomaly score $s_{i,q}$ to every cell using the residual scores defined in \S\ref{sec:anomaly-detection}. We then evaluate whether these scores rank injected cells above non-injected cells.

AUROC is computed as the probability that a randomly selected injected cell receives a higher score than a randomly selected non-injected cell, with ties counted halfway:
\[
\mathrm{AUROC}
=
\Pr(s^+ > s^-)
+
\tfrac{1}{2}\Pr(s^+ = s^-),
\]
where $s^+$ and $s^-$ denote scores from injected and non-injected cells, respectively.

AUPRC is the area under the precision--recall curve obtained by thresholding the anomaly scores. For a threshold $t$, precision and recall are
\[
\mathrm{Prec}(t)=
\frac{\sum_{i,q}\mathbf{1}[s_{i,q}\ge t]m_{i,q}}
{\sum_{i,q}\mathbf{1}[s_{i,q}\ge t]},
\]
\[
\mathrm{Rec}(t)=
\frac{\sum_{i,q}\mathbf{1}[s_{i,q}\ge t]m_{i,q}}
{\sum_{i,q}m_{i,q}}.
\]
We compute the precision--recall curve over all score thresholds and report the area under this curve. Both AUROC and AUPRC are computed at the cell level and averaged over random injection seeds for each injection rate and modality.

\section{Examples}
\label{appendix:examples}

\paragraph{Item-Profile Gallery.} Table~\ref{tab:item_profiles} walks the four corners of the (difficulty $b_q$, sharpness $a_q$) plane on coding-domain static items. The example items are drawn directly from the full-data \textsc{DualEval} fit; pass counts are observed across the 18 evaluated models. This gallery illustrates qualitatively what the Fisher subset experiment in §4.2 measures quantitatively: ranking signal lives in items where both $a_q$ is high and the pass rate is between the extremes, and items in the saturated and uniformly hard corners contribute essentially no ranking information.
  
\begin{table*}[t]
\centering
\footnotesize
\setlength{\tabcolsep}{4pt}
\renewcommand{\arraystretch}{1.15}
\begin{tabular}{@{}p{0.76\linewidth}@{\hspace{4pt}}r@{}}
\toprule
\textbf{Item} & \textbf{Pass\,/\,$a_q$} \\
\midrule
\textbf{A. Frontier-discriminative.} LCB-v6 ``Stone XOR''.
\newline ``There are $N$ bags labeled $1,\ldots,N$. Bag $i$ contains $A_i$ stones. Choose
two bags $A$, $B$, and move all stones from $A$ to $B$\ldots''
\newline \emph{Hard but very sharp ($b\!=\!0.36$, $a_q\!=\!72.7$). Only 5 of 18 models pass;
the joint fit identifies them as systematically the strongest --- large Fisher information
for ranking the frontier.}
& \shortstack[r]{$5/18$\\$a_q\!=\!72.7$} \\
\midrule
\textbf{B. Uniformly hard.} LCB-v6 ``Min of Restricted Sum''.
\newline ``You are given integers $N$, $M$ and three integer sequences of length $M$\ldots''
\newline \emph{Hardest tail of the difficulty distribution ($b\!=\!1.57$); 0 of 18 models
pass. The item provides no ranking signal --- no variation in outcomes to attribute to
ability differences.}
& \shortstack[r]{$0/18$\\$a_q\!=\!4.4$} \\
\midrule
\textbf{C. Borderline-discriminative.} LCB-v6 ``Unique 3-Digit Even Numbers''.
\newline ``You are given an array of digits. Determine the number of distinct three-digit
even numbers that can be formed using these digits\ldots''
\newline \emph{Easy ($b\!=\!-0.48$) but sharp ($a_q\!=\!31.2$). 16 of 18 models pass; the 2
misses are the two weakest. Informative for separating trailing models, not the frontier.}
& \shortstack[r]{$16/18$\\$a_q\!=\!31.2$} \\
\midrule
\textbf{D. Saturated.} MBPP-Plus ``left insertion point''.
\newline ``Write a Python function to locate the left insertion point for a specified value
in sorted order\ldots''
\newline \emph{Trivially easy ($b\!=\!-1.44$), 18 of 18 models pass. Contributes nothing to
ranking; a candidate for retirement under benchmark refresh.}
& \shortstack[r]{$18/18$\\$a_q\!=\!5.0$} \\
\bottomrule
\end{tabular}
\caption{Item-profile gallery (coding domain) walking the four corners of the (difficulty
$b_q$, discrimination $a_q$) plane. Pass counts are out of the 18 evaluated models.
Frontier-discriminative items (A) carry the bulk of ranking signal; uniformly hard (B) and
saturated (D) items contribute none and are candidates for retirement;
borderline-discriminative items (C) inform only the low-ability tail. The Fisher-information
subset experiment (Fig.~\ref{fig:sample-efficiency}) operationalizes this picture
quantitatively.}
\label{tab:item_profiles}
\end{table*}

\paragraph{Anomaly Candidates.} Table~\ref{tab:anomaly_examples} lists representative anomaly candidates surfaced by the standardized \textsc{DualEval} residual diagnostic of \S\ref{sec:anomaly-detection}, restricted to natural (un-perturbed) responses with $|r_{i,q}| > 3.5\sigma$. These are candidates, not verified contamination: large residuals can also reflect MC guessing on items with few options, judge artifacts, or genuine but narrow capability not captured by the joint fit. Semantic-preserving perturbation audits are the appropriate follow-up to distinguish memorization from robust capability and are left to future work. The patterns observed across the three picks (number-sequence puzzle archives, specialized MC questions, named MBPP function signatures) are nevertheless consistent with prior exposure as a plausible alternative hypothesis.
  
\begin{table*}[t]
  \centering
  \footnotesize
  \setlength{\tabcolsep}{4pt}
  \renewcommand{\arraystretch}{1.15}
  \begin{tabular}{@{}p{0.76\linewidth}@{\hspace{4pt}}r@{}}
  \toprule
  \textbf{Anomaly candidate} & \textbf{$\hat p\,/\,r$} \\
  \midrule
  \textbf{HLE-Math} \,/\, \texttt{gpt-4.1} (rank 16/18).
  \newline ``2, 11, 23, 51, 119, ( )\ A.\,291\ B.\,285\ C.\,171\ D.\,167.''
  \newline \emph{MC sequence-completion item; the pattern is widely archived (puzzle
  banks/OEIS) and a weak math model recovering the correct option is consistent with prior
  exposure.}
  & \shortstack[r]{$0.04$\\$+4.81$} \\
  \midrule
  \textbf{HLE-Bio/Med} \,/\, \texttt{llama-4-maverick-instruct} (18/18).
  \newline ``Which genetic disorder caused by mutations on chromosome 2 leads to the greatest
  increases to patients' basal metabolic rate? \ A.\,Alstr\"om\ B.\,Menkes\ldots''
  \newline \emph{Specialized medical MC. The lowest-ranked model passing where frontier models
   fail is $>\!5\sigma$; consistent with recall from medical-Q\&A corpora.}
  & \shortstack[r]{$0.03$\\$+5.36$} \\
  \midrule
  \textbf{MBPP-Plus} \,/\, \texttt{llama-4-maverick-instruct} (18/18).
  \newline ``Write a python function to find the sum of the per-digit difference between two
  integers. Your function must be named \texttt{digit\_distance\_nums}.''
  \newline \emph{MBPP is among the most-trained-on coding benchmarks; the named signature
  suggests recall rather than synthesis from the spec.}
  & \shortstack[r]{$0.07$\\$+3.57$} \\
  \bottomrule
  \end{tabular}
  \caption{Anomaly candidates surfaced by standardized \textsc{DualEval} residuals $r_{i,q} =
  (y_{i,q} - \hat p_{i,q}) / \sqrt{\hat p_{i,q}(1-\hat p_{i,q})}$ on un-perturbed responses
  ($y_{i,q}=1$ for all rows).}
  \label{tab:anomaly_examples}
  \end{table*}




\end{document}